\documentclass[10 pt, conference]{IEEEtran}  

\IEEEoverridecommandlockouts                              

\usepackage{cite} 
\usepackage{amsfonts}
\usepackage{algpseudocode}
\usepackage{algorithm}
\usepackage{hyperref}
\usepackage{float}
\usepackage{graphicx}
\usepackage{textcomp}
\usepackage{svg}
\usepackage{color} 
\usepackage{array}
\usepackage{booktabs}
\usepackage{pifont}
\usepackage{multirow}
\usepackage{multicol}
\usepackage[dvipsnames]{xcolor}
\usepackage{placeins}
\usepackage{gensymb}

\usepackage{dblfloatfix}    
\usepackage{kantlipsum}     

\usepackage{mathptmx} 
\usepackage{times} 
\usepackage{amsmath} 
\usepackage{amssymb}  

\usepackage[lofdepth,lotdepth]{subfig}
\usepackage{booktabs}
\usepackage{graphicx}
\usepackage[dvipsnames]{xcolor}
\usepackage{array}
\usepackage{caption}
\usepackage{comment}

\newcolumntype{M}[1]{>{\centering\arraybackslash}m{#1}}

\usepackage[
  left=0.667in,
  right=0.667in,
  top=0.833in,
  bottom=0.597in
]{geometry}
\usepackage{hyperref}
\hypersetup{
    colorlinks=true,
    linkcolor=blue,
    filecolor=magenta,
    citecolor=teal,
    urlcolor=red,
    pdftitle={Overleaf Example},
    pdfpagemode=FullScreen,
    }

\usepackage[normalem]{ulem} 

\title{\LARGE \bf
Are We Ready for Snow-tonomous Driving? \\ Challenges Toward Robust Automated Driving in Extreme Weather 
}

\author{
Eren Erdal Aksoy$^{1,2}$, Abu Mohammed Raisuddin$^{1}$, Thien-Thu Ngo$^{1}$,   
Eren Aydemir$^{3}$, Osman Tahir Ekşi$^{3}$, \\
Paula  Litkey$^{4}$,  
Petri Manninen$^{4}$, Heikki Hyyti$^{4}$, Jyri Maanpää$^{4}$, Tarmo Kivioja$^{4}$,
Tuomas Herranen$^{5}$, \\ 
Väisänen Sofie$^{5}$, Niemi Kalle$^{5}$, 
Isabelle Morvan$^{6}$,  
Pierre Duthon$^{7}$,
Ian Marsh$^{8}$, Martin Sanfridson$^{8}$,  \\ 
 Henrik Toss$^{8}$,  Thanh Bui$^{8}$, 
Yuri Poledna$^{9}$,  
Maikol Funk Drechsler$^{9}$, 
Mikko Tarkiainen$^{10}$,  
Matti Kutila$^{10}$,  \\
Topi Miekkala$^{10}$,
Stefan Parzefall$^{11}$,
Valentina Donzella$^{12}$, Pak Hung Chan$^{12}$, 
Sogol Kharrazi$^{13}$,
Mario Ceccarelli$^{14}$\\ 
\thanks{$^{1}$Halmstad University, School of Information Technology, Sweden}%
\thanks{$^{2}$Lund University, Department of Computer Science, Sweden.}
\thanks{$^{3}$FORD Otosan, Turkey.}
\thanks{$^{4}$Finnish Geospatial Research Institute,  Finland.}
\thanks{$^{5}$Lapland University of Applied Sciences, Finland.}
\thanks{$^{6}$CRF, France.}
\thanks{$^{7}$CEREMA, France.}
\thanks{$^{8}$RISE, Sweden.}
\thanks{$^{9}$CARISSMA Institute of Automated Driving, TU Ingolstadt, Germany.}
\thanks{$^{10}$VTT, Finland.}
\thanks{$^{11}$AVL, Germany.}
\thanks{$^{12}$Queen Mary University of London, United Kingdom.}
\thanks{$^{13}$VTI, Swedish National Road and Transport Research Institute, Linkoping, Sweden.}
\thanks{$^{14}$accelopment Schweiz AG, Switzerland.}
}

\begin{document}
\makeatletter
\setcounter{figure}{-2}
\let\@oldmaketitle\@maketitle
\renewcommand{\@maketitle}{\@oldmaketitle
\centering 
    \includegraphics[width=1\textwidth]{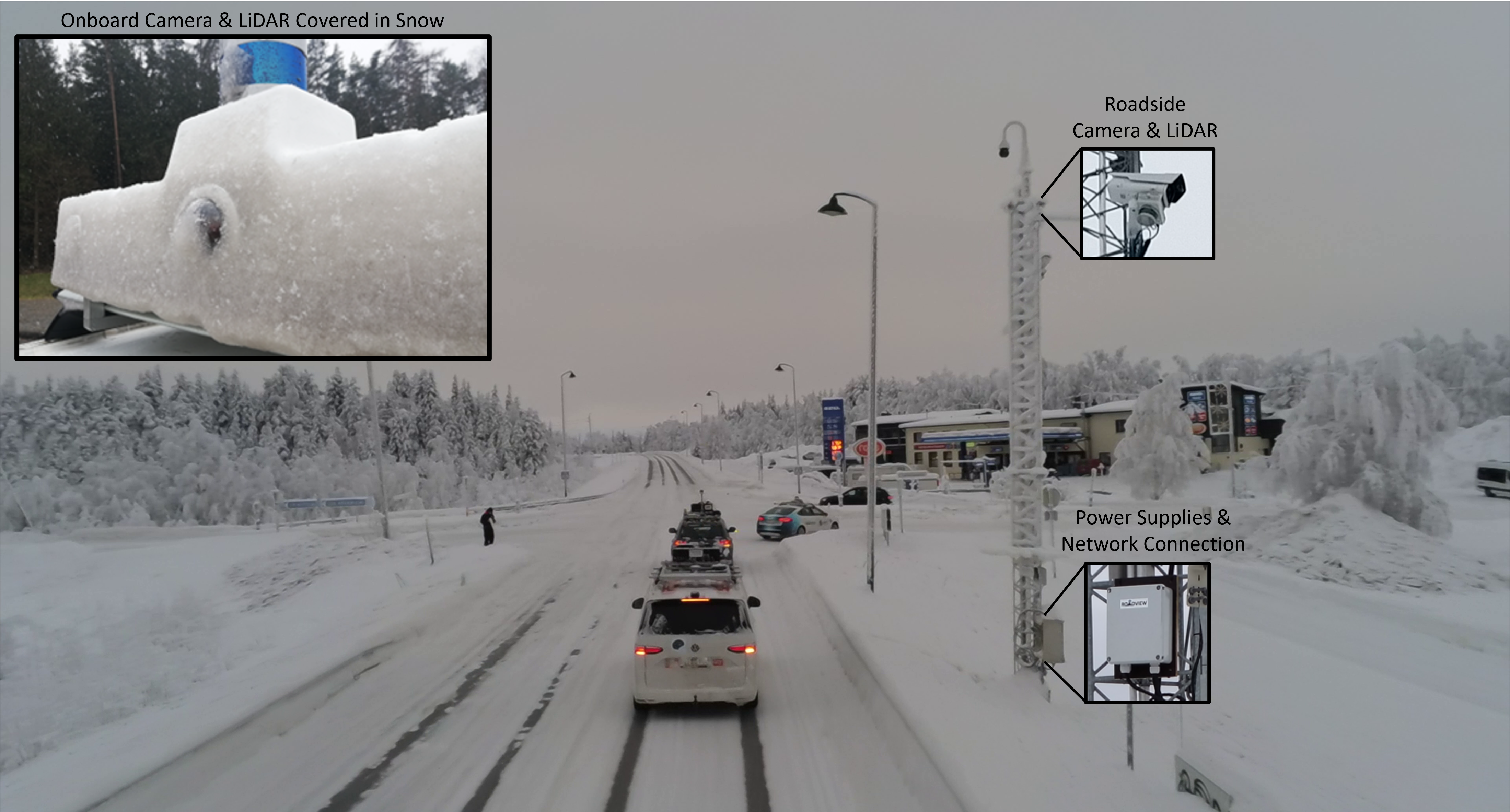}
   \captionof{figure}{ROADVIEW demonstration scene in heavy snowy conditions in Finland, where the minimum recorded temperature during testing was -34$\degree$C (Click \href{https://www.youtube.com/watch?v=4X_0BrStTMw}{here} to see the demonstration video). The top-left image shows the onboard camera and LiDAR sensors from a test vehicle, both heavily covered with snow and ice.  
   }
    \label{fig:intro_fig} 
   }


\maketitle

\thispagestyle{empty}
\pagestyle{empty}
\begin{abstract}
Automated driving has gained remarkable momentum over the past decade, with both industry and academia investing substantial resources in developing perception and control solutions to advance vehicle autonomy.
However, these efforts have largely focused on clear-weather scenarios, leaving harsh conditions, such as snow, rain, and fog, comparatively understudied. 
As a result, inclement weather remains one of the most significant hurdles for autonomous vehicles,
often leading to substantial performance degradation in downstream perception and control tasks.

In this paper, we present insights from the EU-funded Horizon Europe project ROADVIEW, which aims to develop robust, weather-aware perception and control modules for connected and automated vehicles operating under severe conditions, including snow, rain, and fog. Drawing on four years of large-scale development and real-world demonstrations across multiple vehicle platforms in different locations, we identify ten key challenges that hinder the development of weather-resilient autonomy, including data scarcity, domain shifts, limitations in multimodal perception, control robustness, and validation gaps, among others. 
We summarize these challenges and lessons learned toward enabling reliable \textit{snow-autonomous} driving, highlighting critical directions for future research and standardization.


\end{abstract}
 
%
%
\section{Introduction}
\label{Sec:Introduction}

Over the past decade, automated driving research has undergone a quantum leap, largely driven by rapid and exciting advances in Artificial Intelligence (AI). Both academic research institutions and companies have invested substantial resources in developing and demonstrating AI-based perception and control systems across increasingly impressive real-world scenarios. Notable examples include Waymo’s autonomous robotaxi fleet\footnote{\url{https://waymo.com/}} operating continuously in the US and the large-scale deployment of Baidu’s robotaxis\footnote{\url{https://ir.baidu.com/}} across multiple cities in China~\cite{esteve2026smart}.
Despite these remarkable achievements, a fundamental question remains: \textit{why do we not yet see robotaxis navigating autonomously in harsh weather (e.g., under heavy rain or snowfall)?} 

This breakdown in autonomy under extreme environmental conditions highlights a lack of systemic resilience. 
The challenge is particularly pressing considering the fact that, for instance, in March 2025 alone, nearly 3 million km$^2$ of Europe was covered by snow~\cite{esotc2025}.
Thus, until Autonomous Vehicles (AVs) can robustly perceive and safely act in such adverse, unpredictable conditions, the vision of truly reliable autonomy remains unfulfilled.

Severe weather is a natural phenomenon that poses significant challenges for current AVs. 
Sensors can become partially (or fully) occluded or contaminated by water droplets, mud, snow, and ice (see Fig.~\ref{fig:intro_fig}), leading to degraded sensor measurements and impaired perception performance. 
For example, 3D LiDAR scans can be severely affected by precipitation, which in turn degrades downstream tasks such as object detection and semantic segmentation.
Likewise, road surface conditions can deteriorate rapidly, for instance, due to ice formation or aquaplaning. This compromises tire–road friction and reduces controller reliability and safety. Testing and validation under such conditions remain particularly challenging, as severe weather is both sporadic and difficult to reproduce in a controlled and scalable manner.

To address these weather-related challenges, the EU-funded Horizon Europe project ROADVIEW\footnote{\url{https://roadview-project.eu/}} (RObust Automated DriVIng in Extreme Weather) aims to develop robust, weather-aware perception and control modules that enhance the performance of connected and automated vehicles under severe conditions, including snow, rain, and fog (See Fig.~\ref{fig:intro_fig}). 
With 15 partners from seven countries, ROADVIEW targets a Technology Readiness Level (TRL) 7 automated driving stack deployed across five distinct vehicle platforms, including passenger cars and trucks (See Figs.~\ref{fig:intro_fig},~\ref{fig:multi_domain}, and~\ref{fig:denoising}).
Building on four years of intensive development and real-world deployment in ROADVIEW, we pose a fundamental question to the research community: \textit{are we truly ready for \textit{snow-autonomous} driving?}

Drawing on lessons learned from multiple ROADVIEW demonstrations, we, as members of the ROADVIEW consortium, identify \textbf{ten core challenges} as key obstacles to achieving autonomy resilient to weather-induced disturbances. These challenges are summarized as follows:

\begin{enumerate} 
    \item \textbf{Lack of annotated harsh weather datasets}: 
    Although several autonomous driving datasets~\cite{mantruckscene,kent2024msu,xiao2023,zod2023,Ithaca365,pitropov2021canadian,kurup2021dsor} include all-weather scenarios (see Table~\ref{tab:dataset_comp}), most contain only a limited amount of annotated data representing severe weather conditions. For example, snowy scenes account for merely $2\%$ of the ZOD~\cite{zod2023} dataset and approximately $12\%$ of Ithaca365~\cite{Ithaca365}. Some datasets are specifically designed for winter environments~\cite{pitropov2021canadian,xiao2023}; however, they typically focus on a single perception task, such as 3D object detection in the CADC dataset~\cite{pitropov2021canadian} or 3D semantic segmentation in SemanticSTF~\cite{xiao2023}. Furthermore, existing publicly available datasets lack comprehensive ground-truth annotations for jointly evaluating multiple multimodal 3D perception tasks under harsh winter conditions. Thus, the research community is in urgent need of multimodal datasets that span multiple seasonal conditions, ensure balanced data distributions, and provide supervision signals for multiple perception tasks.


    \item \textbf{Multimodal sensing and data quality}:
    Sensing is a fundamental component for bootstrapping autonomy. 
    Severe weather, however, introduces dynamic and modality-specific degradations, such as reduced visibility in RGB images, snow-induced noise in LiDAR point clouds, and signal ambiguities in RADAR data.
    The challenge arises when noise in raw 2D images and 3D point clouds propagates to downstream perception tasks, leading to degraded performance. 
    Contemporary research tackles these heterogeneous disturbances that vary over time and space by applying conventional statistical noise filtering techniques~\cite{rusu20113d, charron2018noising, kurup2021dsor, park2020fast, huang2023lidsor} or advanced AI-based supervised denoising methods~\cite{seppanen20224denoisenet,heinzler2020cnn, raisuddin20243d}, which are unimodal and require substantial amounts of annotated data. 
    Still, the development of sensor-agnostic and generic filtering methods remains largely underexplored.
    Furthermore, there is no established consensus on a metric to quantify data quality levels, akin to Technology Readiness Levels (TRLs)~\cite{annex_trl}. 
    Therefore, multimodal data quality degradation and sensor impairments need to be assessed from sensor installation at the hardware level to object detection at the machine learning level~\cite{Marsh:IV20265, BuiMarsh:ICHCAI2026}. 
    
    
    \item \textbf{Real-time weather-aware multimodal perception}: 
    While multimodal sensing provides complementary information and enhances robustness through increased redundancy, achieving real-time perception while continuously adapting to changing weather conditions remains a significant challenge.
    A key limitation of current approaches is that they remain context-agnostic and follow static fusion strategies, in which the relative contributions of each modality are fixed and fail to adapt to variations in scene context, such as weather-induced degradation.  
    Although recent methods explore attention-based cross-modal fusion \cite{bai2022transfusion} and availability-aware weighting schemes \cite{paek2026availabilityaware}, their applicability in real-time settings is often limited by computational complexity.
    Consequently, balancing robustness, adaptability, and computational efficiency remains an open challenge for ensuring safe and reliable autonomy.  
    To further enhance resilience against harsh weather conditions, more advanced perception solutions are required, for instance, to distinguish various weather types, estimate road-surface slipperiness, and predict the visibility of the drivable area, among others. 
 
    
    \item \textbf{Weather-aware map-based positioning}:
    Accurate vehicle positioning relative to the surrounding roadway is fundamental for AVs. In addition to satellite-based positioning, High-Definition (HD) map-based positioning is widely used to estimate the vehicle’s pose with respect to a prior map of the environment, enabling centimeter-level accuracy and lane-level semantic understanding~\cite{elghazaly2023high}.
    HD maps commonly use a set of points and line segments in a global reference frame to represent detectable features in the environment. 
    These features include lane markings, road boundaries, lane dividers, curbs, sidewalks, pedestrian crossings, and both vertical and horizontal traffic signs.
    Owing to their rich and precise representation of the road environment, HD maps are considered among the most suitable priors for localization~\cite{javanmardi2018factors}. 
    However, many of these features become partially or fully obscured under harsh winter conditions, when snow obscures most road markings.
    Positioning the automated vehicle within the required accuracy of 10 cm  \cite{reid2019localization} under these conditions is therefore challenging and requires novel, robust methods.

    \item \textbf{Collective perception via V2X communication}:
    Collective perception enables vehicles and roadside units to extend their understanding of the environment by sharing high-level perception results. Current V2X~\cite{v2x} deployments remain limited to connected-vehicle self-announcements, which provide insufficient situational awareness, especially at low market penetration or in the presence of non-connected road users, such as pedestrians or animals. The next generation of V2X services introduces the Collective Perception Message (CPM) to share the outputs of a single sensor or a multimodal perception system, including sensor metadata, perception regions, and perceived objects. Within ROADVIEW, we investigated how adverse weather affects the quality of perception systems and which quality-related information should be shared in CPMs \cite{morvan_sw_2026}. To support this work, an infrastructure-based perception system was installed in Northern Finland to evaluate the impact of harsh weather conditions on the sensing performance. We also explored methods \cite{cancouet_mitigating_2026} to mitigate systematic errors in roadside sensing by performing offline deviation analysis with a high-precision probe vehicle.
    The integration of V2X-based collective perception into automated driving requires reliable temporal and spatial alignment between local perception and CPMs received from roadside units or other vehicles. This alignment depends on the availability of uncertainty descriptors, detection confidence, and kinematic precision to support robust fusion.
    Early deployment phases will involve only a limited number of CPM-capable vehicles or roadside units, but the system must remain scalable as penetration increases. To avoid radio channel congestion, value-of-information-based mechanisms \cite{rolich_trade-off_2025} are needed to dynamically adapt CPM transmission frequency and message size.
    Cross-vendor interoperability also remains a significant challenge: standardization activities for CPMs are still evolving, and additional profiling work is required to ensure consistent implementation across manufacturers, including recommended parameter settings and harmonized interpretation of quality fields.
    These lessons highlight that large-scale deployment of collective perception depends on both high-quality perception outputs and continued progress in standardization and profiling activities.


    \item \textbf{Weather-aware decision making}:
    Adverse weather affects automated driving not only by degrading perception but also by altering the vehicle's safe operating envelope. Reduced visibility, sensor degradation, heavy snowfall or rain, fog, and low tyre–road friction can increase braking distance, reduce maneuverability, and make safe speed selection dependent on uncertain environmental estimates. Prior work has studied friction-aware vehicle control \cite{santini2021board} and LiDAR-based visibility estimation in fog \cite{miclea2020laser}, \cite{montalban2023bayesian}, but a key unresolved challenge remains: how to translate these uncertain, spatially variable estimates of road and weather conditions into reliable driving decisions. This includes determining when the vehicle should reduce speed, increase safety margins, restrict maneuvers, or trigger a minimum-risk maneuver. The difficulty is that weather-related estimates cannot be directly converted into control commands; they must be interpreted alongside vehicle dynamics, route geometry, ODD limits, and uncertainty to ensure safe automated driving under changing conditions. 

    
    \item \textbf {Sim-to-real gap in noise modeling}:
    In AI-based driving systems that rely heavily on sensor data, it is essential to simulate and emulate large volumes of data for training, validation, and verification. However, since driving functions are safety-critical, collecting sufficient real-world data to cover extreme conditions and complex scenarios is practically infeasible. Within ROADVIEW, significant effort has been devoted to sensor data generation and augmentation to enable thorough system testing under conditions where data cannot be collected or controlled, with a particular focus on adverse weather, including snow, rain, and fog.
   
    Nevertheless, the use of such synthetic or augmented data must be accompanied by a rigorous analysis of its validity for the specific testing scope and environment (see the following subsection), complemented by validation against real-world data. This validation is necessary to assess the Sim2Real (Simulation-to-Real) gap. In this context, ROADVIEW has evaluated the impact of weather-induced noise models (i.e., rain, fog, and snow) on camera, LiDAR, and 4D RADAR data, considering both data-driven and physics-based approaches \cite{chan2024parametric, rvD33,rvD34,D3.3Code,D3.4Code,rvD35, Duthon2024a, Toss2025}. The key lessons learned are summarized as follows:

    \textbf{i. Importance and controllability of real data for validation.} 
    Noise models need to be evaluated against real data collected in a controlled manner (e.g., with gradually increasing snowfall intensity) and, ideally, include `paired frames', meaning identical scenes captured both with and without noise under the same conditions. In practice, achieving this is extremely challenging, as even minimal variations in environmental factors (e.g., wind or illumination) can significantly affect the captured data, even in controlled outdoor setups. To address this challenge, ROADVIEW developed and curated the REHEARSE dataset \cite{REHEARSE2024} with noise annotations \cite{REHEARSE-3D}.
  
    \textbf{ii. Lack of standardized metrics.} There is currently no widely accepted evaluation framework in this domain. Within ROADVIEW, we propose a structured methodology based on point cloud metrics \cite{shahbeigi2024novel} and image quality metrics \cite{gummadi2023correlating} to assess the Sim2Real gap. However, this approach provides only a partial evaluation, and further work is needed to develop holistic methods for assessing the Sim2Real gap in generated data and digital twins.
    
    \textbf{iii. Purpose-driven fidelity.} 
    Even when highly accurate models are available, we recommend that future Sim2Real gap evaluations be linked to downstream perception performance of the overall automated driving pipeline \cite{donzella2025lidar}. Generating extremely realistic data is not inherently beneficial if system performance is not sensitive to such fidelity. Instead, it is crucial to understand how AI-extracted features relate to data artifacts and representations, and to focus evaluation efforts on those aspects that meaningfully affect system behavior.

    \item \textbf{XiL testing and validation}:
    XiL (X-in-the-Loop) testing systems enable early verification and validation of automotive software through virtual environments \cite{ISO26262.2018}.
    At advanced levels of integration, where the software is deployed on the target hardware (Hardware-in-the-Loop), real-time simulation constraints become essential to guarantee reproducible and accurate evaluations \cite{Drechsler_2022_Simulation}.     
    The integration of complex models, such as weather models, introduces a trade-off between fidelity and computational cost, often limiting their applicability to a restricted set of weather conditions. 
    Meaningful and generalizable applications require deep integration into simulation engines, which, in turn, makes models tightly coupled to specific simulation environments and hinders interoperability across simulators. 
    Moreover, the lack of standardized methods for accessing simulation physics and for applying sensor manipulation remains a critical barrier to the broader adoption of XiL systems.     

    Despite recent initiatives to create databases of LiDAR reflectivity\footnote{https://www.persival.de/material-database}, material properties are still not comprehensively cataloged. 
    Furthermore, current standards do not support the dynamic adaptation of these properties under varying weather conditions. 
    For example, in the modeling of road slipperiness, existing standards do not allow the definition of spatially varying or dynamic slippery zones, thereby limiting testing under realistic, changing conditions\footnote{https://www.asam.net/standards/detail/opencrg/}. 
    Similarly, standard models available in open-source simulators are often simplified \cite{CARLA} and limited in their representation of real-world complexity~\cite{Ulreich.2025}.
    The lack of high-fidelity, low-latency sensor models continues to constrain the application of XiL systems, hindering the development and validation of automated systems not only in the automotive domain but across the broader robotics ecosystem.
   
    Recent EU regulations explicitly encourage the use of simulation for AV validation, emphasizing the need to validate simulation results against proving-ground and real-world testing \cite{EU2019_2144, UNR157_2021, ECE2021}. 
    However, methodologies for validating simulation credibility are not yet fully defined, and the extrapolation of validation results to real-world scenarios remains insufficiently addressed.
    Furthermore, reliance on proprietary interfaces for data exchange between sensors and control units limits the ability to perform standardized assessments and homologation of production vehicles using sensor data injection. 
    The industry, therefore, requires standardized interfaces and clear validation methodologies to ensure the credibility of simulations, as their absence impedes the large-scale adoption of XiL approaches for AVs and delays their deployment.

    \item \textbf{Transparency in perception and control}:
    A key challenge in AI-based perception and control is the lack of transparency in how AI models transform sensor inputs into predictions. These systems often operate as opaque black boxes~\cite {Bathaee2018TheAI}, making it difficult for human experts to assess whether their predictions rely on relevant, sufficient, and meaningful input features. Such opacity poses a fundamental challenge for the verification and validation of AI-based systems against safety requirements, where reliable performance shall be guaranteed across diverse ODD conditions, including harsh weather. Without sufficient transparency and traceability, it is challenging to ensure system safety during operation.
    
    Another challenge lies in establishing trustworthiness when perception outputs are directly consumed by downstream control modules for safety-critical decisions~\cite{rob-11015007}. 
    For example, a weather-aware control module may rely on perception-based estimates of weather conditions and visibility. In such cases, quantifying uncertainty and detecting out-of-distribution scenarios are essential to prevent overconfident predictions from propagating into unsafe actions in severe weather~\cite{BuiMarsh:ICHCAI2026}.

    \item \textbf{Standardizations and regulation}: 
    Despite their potential, the large-scale commercialization and deployment of weather-resilient AV perception and control solutions remain constrained by regulatory and policy barriers. To fully unlock their benefits, policymakers (e.g., the EU and its Member States) must align fragmented connected and automated vehicle regulatory frameworks, clarify liability regimes, and establish harmonized and standardized testing and validation methodologies for adverse weather conditions.
    In addition, clear and enforceable guidelines on AI transparency, data governance, privacy, and V2X communication standards are essential to enable scalable deployment while preserving public trust and ensuring regulatory compliance.
    Policymakers should actively foster cross-border collaboration, invest in resilient transnational data-sharing infrastructures, and streamline regulatory processes to support consistent and safe deployment.
     
\end{enumerate}


\section{Challenges}
\label{sec:challenges}
This section details the ten challenges, supported by quantitative and qualitative evidence. For each of these challenges, we also provide our recommendations based on lessons learned from ROADVIEW.

\subsection{Lack of annotated harsh weather datasets}
\label{sec:sec_dataset}
 
\begin{table*}[!t]
\caption{Comparison of SoA outdoor public autonomous vehicle datasets. BB, SL, and SF refer to 3D annotations for Bounding Boxes, Semantic Labels, and Snow Flakes, respectively.   
- represents unsupported or no statistics.
}
\label{tab:dataset_comp}
\begin{center}
\resizebox{1\textwidth}{!}{
\begin{tabular}{lcccc ccccc cccc}
\toprule 
\multirow{2}{*}{} & & \multicolumn{5}{c}{\bfseries{Modality}} & \multicolumn{3}{c}{\bfseries{3D Annotation}} & & & \\ \cmidrule(lr){3-7}\cmidrule(lr){8-10}
\bfseries{Dataset} & \bfseries{Year} & \bfseries{LiDAR} & \bfseries{RGB} & \bfseries{Radar} & \bfseries{Thermal} & \bfseries{GNSS/IMU} &
\bfseries{BB} & \bfseries{SL} & \bfseries{SF} & \bfseries{\#Frames} & \bfseries{\#Classes} & \bfseries{Coverage} & \bfseries{Snow}\\ \toprule
KITTI~\cite{geiger2012we}   & 2012 & 64 & 90$^{\circ}$  & - & - & \ding{51}/\ding{51}  & \ding{51} &- & -& 15K & 8 & 39.2 km &- \\ 
KAIST~\cite{KAIST_dataset}  & 2018 & 32 & 26$^{\circ}$  & - & 25$^{\circ}$ & \ding{51}/\ding{51} &\ding{51} & - & -& - & 3 & 44.8 km &-\\ 
nuScenes~\cite{caesar2020nuscenes}  & 2019 & 32 & 360$^{\circ}$   & \ding{51} & - &  \ding{51}/\ding{51} & \ding{51} & \ding{51}  & - & 40K & 23 & 242 km &- \\ 
Waymo~\cite{waymo_dataset}   & 2020 & 64 & 360$^{\circ}$  & - & - & \ding{51}/\ding{51}  & \ding{51} & \ding{51}  & - & 230K & 4 & 1750 km &-\\ 
A2D2~\cite{a2d2dataset}  & 2020 & 16 & 360$^{\circ}$   & - & - &  \ding{51}/\ding{51}   &  \ding{51} & -  & - & 12K & 14 & - &-\\ 
Argoverse 2~\cite{Argoverse2}   & 2021 & 32 & 360$^{\circ}$  & - & -  & - / -   & \ding{51} & -  & - & 150K & 30 & 2000 km  &-\\ 
SemanticKITTI~\cite{behley2021ijrr}  & 2021 & 64 & -   & - & - & \ding{51}/\ding{51}   & -& \ding{51}  & -& 43K & 28 & 39.2 km &-\\ 

\midrule
DENSE~\cite{bijelic2020seeing}  & 2019 & 64 & 120$^{\circ}$  & - & 57$^{\circ}$ &  - / -  & \ding{51} & -  & -  & 12K & 5 & - & \ding{51}\\ 
CADC~\cite{pitropov2021canadian}  & 2021 & 32 & 360$^{\circ}$    & -  & - & \ding{51}/\ding{51} & \ding{51}& -  & - & 7K & 10 & 20 km & \ding{51}\\ 
WADS~\cite{kurup2021dsor} & 2021 & 64 & -  & -  & - & - / - & - &  \ding{51} &  \ding{51}  & 2K & 22 & - & \ding{51}\\ 
Ithaca365~\cite{Ithaca365}   & 2022 & 128 & 60$^{\circ}$   & -  & - & \ding{51}/\ding{51} & \ding{51} & -  & - & 7K & 6 & 15 km & \ding{51}\\ 
Boreas~\cite{Boreas}  & 2023 & 128 & 81$^{\circ}$   & \ding{51}  & -  &  \ding{51}/\ding{51} & \ding{51}  & - & -& 7K & 3 & 350 km & \ding{51}  \\ 
ZOD~\cite{zod2023}  & 2023 & 128 & 120$^{\circ}$    & - & - &  \ding{51}/\ding{51} & \ding{51}  & - & -& 100K & 29 & - & \ding{51} \\ 
SemanticSTF~\cite{xiao2023}  & 2023 & 64 & - & - & - & - / -& - &  \ding{51}  & -& 2K & 21 & - & \ding{51} \\ 
MSU-4S~\cite{kent2024msu}  & 2024 & 64 & 150$^{\circ}$   & \ding{51}  & - & \ding{51}/\ding{51}  & \ding{51} & -  & - & 100K & 3 & 14.8 km  & \ding{51} \\ 
MAN TruckScenes~\cite{mantruckscene}  & 2024 & 64 & 360$^{\circ}$    & \ding{51} & - & \ding{51}/\ding{51} & \ding{51} &  -  & -& 30K & 27 & - & \ding{51}\\ 
Snowy Scenes~\cite{snowyscenes}    & 2026 & 128 & 70$^{\circ}$ & -  & 72$^{\circ}$ & \ding{51}/\ding{51}  &  \ding{51} &  \ding{51} & \ding{51}   & 5K & 27 & 14.4 km & \ding{51}\\ 
M3 Scenes~\cite{m3scenes}    & 2026 & 16 & 360$^{\circ}$  & \ding{51} & -  & \ding{51}/\ding{51}   &  \ding{51} &  \ding{51} & \ding{51}   & 13K & 6 & 800 km & \ding{51}\\\bottomrule
%
\\
\end{tabular}
}
\end{center}
\end{table*}

Over the past decade, autonomous driving research has made substantial progress, supported by large-scale benchmark datasets such as KITTI~\cite{geiger2012we}, nuScenes~\cite{caesar2020nuscenes}, and Waymo~\cite{waymo_dataset}, to name a few. 
These datasets have played a pivotal role in advancing multimodal perception for urban environments. However, their data distributions are heavily skewed toward clear weather, limiting their applicability in adverse environments, particularly in winter scenarios characterized by snow, low visibility, and challenging illumination conditions.

An overview of the current dataset landscape is shown in Table~\ref{tab:dataset_comp}, where existing datasets are categorized based on the presence of snowy conditions. 
While snow is present in a subset of datasets, particularly those in the bottom half of Table~\ref{tab:dataset_comp}, it often constitutes only a small portion of the overall data.
For instance, only $2\%$ of the ZOD~\cite{zod2023} dataset (about $2$K frames) and $12\%$ of Ithaca365~\cite{Ithaca365} (approximately 800 frames) involve snow.
In SemanticSTF~\cite{xiao2023}, this rate reaches $33.4\%$,  however, this still corresponds to only 700 frames.
Other datasets, including MSU-4S~\cite{kent2024msu} and MAN TruckScenes~\cite{mantruckscene}, provide more winter samples but offer only 3D bounding-box annotations, with limited semantic detail.
Furthermore, in datasets such as Boreas~\cite{Boreas}, 3D annotations are available only for clear-weather conditions, further highlighting the imbalance.

Datasets specifically designed for winter environments remain scarce. 
CADC~\cite{pitropov2021canadian} and WADS~\cite{kurup2021dsor} are notable for focusing exclusively on snowy conditions. CADC introduced a fully winter dataset tailored for 3D object detection, while WADS extended this line of work by incorporating 3D semantic segmentation with explicit snowflake annotations. Despite these contributions, both datasets exhibit important limitations. CADC relies on relatively low-resolution 32-beam LiDAR data, whereas WADS is constrained to a single sensing modality. In addition, both lack diversity in annotation types and multimodal integration, which are critical for comprehensive perception benchmarking.
It is also worth noting that Table~\ref{tab:dataset_comp} excludes several datasets that cover snowy conditions because they lack 3D information. For instance, MUSES~\cite{brodermann2024muses} does not include 3D annotations, BDD100K~\cite{yu2020bdd100k} lacks LiDAR data, and RADIATE~\cite{radiate} provides only 2D RADAR annotations. 

\begin{figure*}[!t]
    \centering
    \includegraphics[width=1\linewidth]{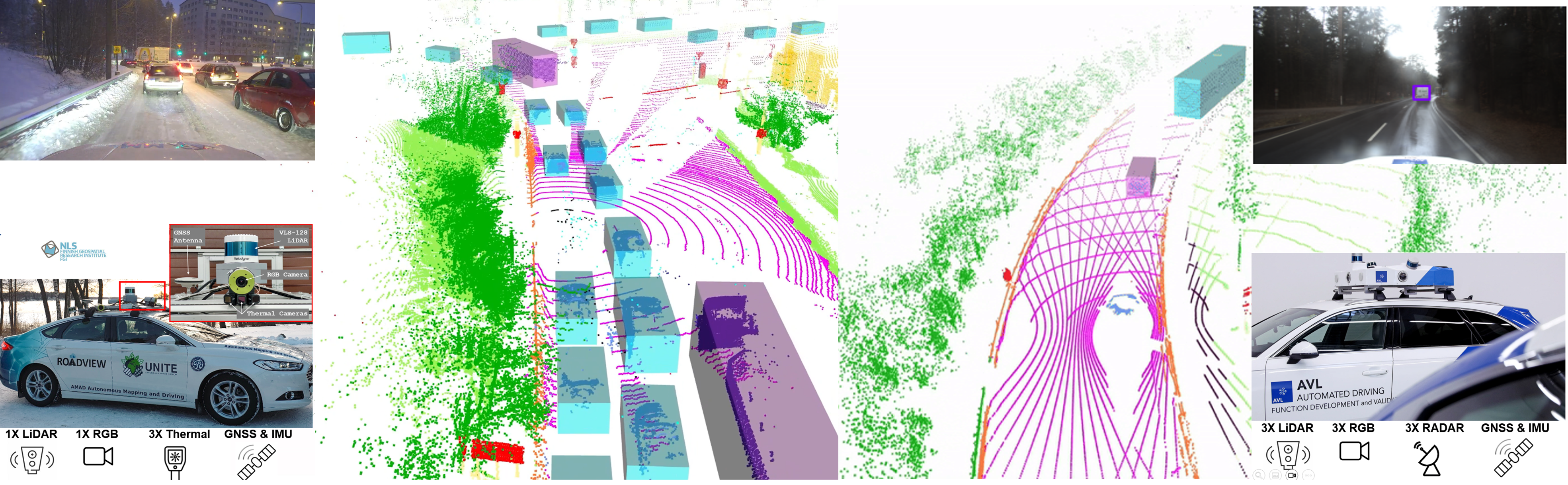}
    \caption{Sample images from the ROADVIEW Snowy Scenes~\cite{snowyscenes} (left) and M3 Scenes~\cite{m3scenes} (right) datasets. Both datasets have different multimodal sensor setups and provide dense point-wise annotations, including falling snowflakes, and 3D bounding boxes to benchmark 3D semantic segmentation, object detection, and denoising under harsh weather conditions.     
    }
    \label{fig:multi_domain}
\end{figure*}

\textbf{Challenge.} Consequently, the current dataset ecosystem shown in Table~\ref{tab:dataset_comp} reveals a significant challenge: \textbf{\textit{a clear lack of datasets that jointly support multiple multimodal 3D perception tasks under harsh winter conditions, with sufficient scale and annotation depth.}} Addressing this challenge is essential to developing robust autonomous driving systems that operate reliably in real-world environments.

\textbf{Recommendation.} As the first attempt in the literature to address this challenge, the ROADVIEW consortium recently introduced two challenging harsh weather datasets, Snowy Scenes~\cite{snowyscenes} and M3 Scenes~\cite{m3scenes} (see the last two rows in Table~\ref{tab:dataset_comp}).   
Snowy Scenes~\cite{snowyscenes} is the first multimodal dataset collected in real-world snowy conditions, using high-resolution 128-beam LiDAR, RGB, and thermal cameras coupled with GNSS/IMU. 
Similarly, M3 Scenes~\cite{m3scenes} spans over 8 hours of continuous driving, covering approximately 800 km across diverse routes, environments, and seasons, and includes  LiDAR, RADAR, and RGB sensors. 
As illustrated in Fig.~\ref{fig:multi_domain}, these ROADVIEW datasets represent a significant advancement over other datasets by providing the most comprehensive set of annotations, enabling research on 3D object detection, semantic segmentation, and point cloud denoising, both as standalone tasks and within joint multitask learning frameworks.

Following our experience in ROADVIEW, we encourage the research community to move beyond single-task datasets with limited weather coverage. Instead, efforts should prioritize (i) the integration of diverse sensing modalities (e.g., LiDAR, RADAR, RGB, thermal, event-based, gated, etc.) to ensure robustness across varying weather conditions, (ii) the provision of rich annotation schemes that enable joint learning of tasks such as 3D object detection, semantic segmentation, and the denoising of adverse weather artifacts, and (iii) the collection of long-duration, geographically diverse data to capture real-world variability. Additionally, we advocate for standardized benchmarks and evaluation protocols that explicitly target multitask performance under harsh weather conditions, thereby fostering a more holistic and realistic assessment of perception systems.
The ROADVIEW datasets represent an initial step in this direction, and we hope they inspire further collaborative efforts toward scalable, richly annotated, multimodal datasets for winter driving scenarios.
%
%
\subsection{Multimodal sensing and data quality}
\label{sec:multimodalsen}

\begin{figure}[!t]
    \centering
    \includegraphics[width=1.0\linewidth]{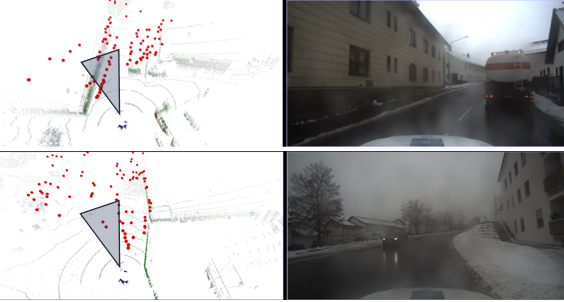}
    \caption{Sample snowy scenes from~\cite{m3scenes} logged at different time steps. The time difference between the top and bottom rows is 8.5 seconds. The left column shows LiDAR point clouds in gray and RADAR data points in red. This difference is also observable in the front-facing camera.}
    \label{fig:snow_time_diff}
\end{figure}

AVs face substantial challenges in perception and navigation, particularly under adverse weather conditions such as rain, fog, and snow. These conditions can severely degrade the performance of key sensing modalities, including LiDAR, RADAR, RGB cameras, and thermal sensors.
Consider two wet and snowy scenes from the ROADVIEW data logging campaigns~\cite{m3scenes} shown in Fig.~\ref{fig:snow_time_diff}, which are separated by 8.5 seconds. The cones in the figure highlight how data points become increasingly sparse over time within the ego vehicle’s forward field of view.
Under snowfall, approaching vehicles become visible in LiDAR point clouds only at close range.
RADAR measurements also degrade due to snowfall. However, the data density remains within a usable range for several more minutes.
Such degradation adversely affects downstream perception tasks, including object detection, semantic segmentation, localization, mapping, and tracking~\cite{rain2018,park2020fast}.

\begin{figure}[!t]
    \centering
    \includegraphics[width=0.7\linewidth]{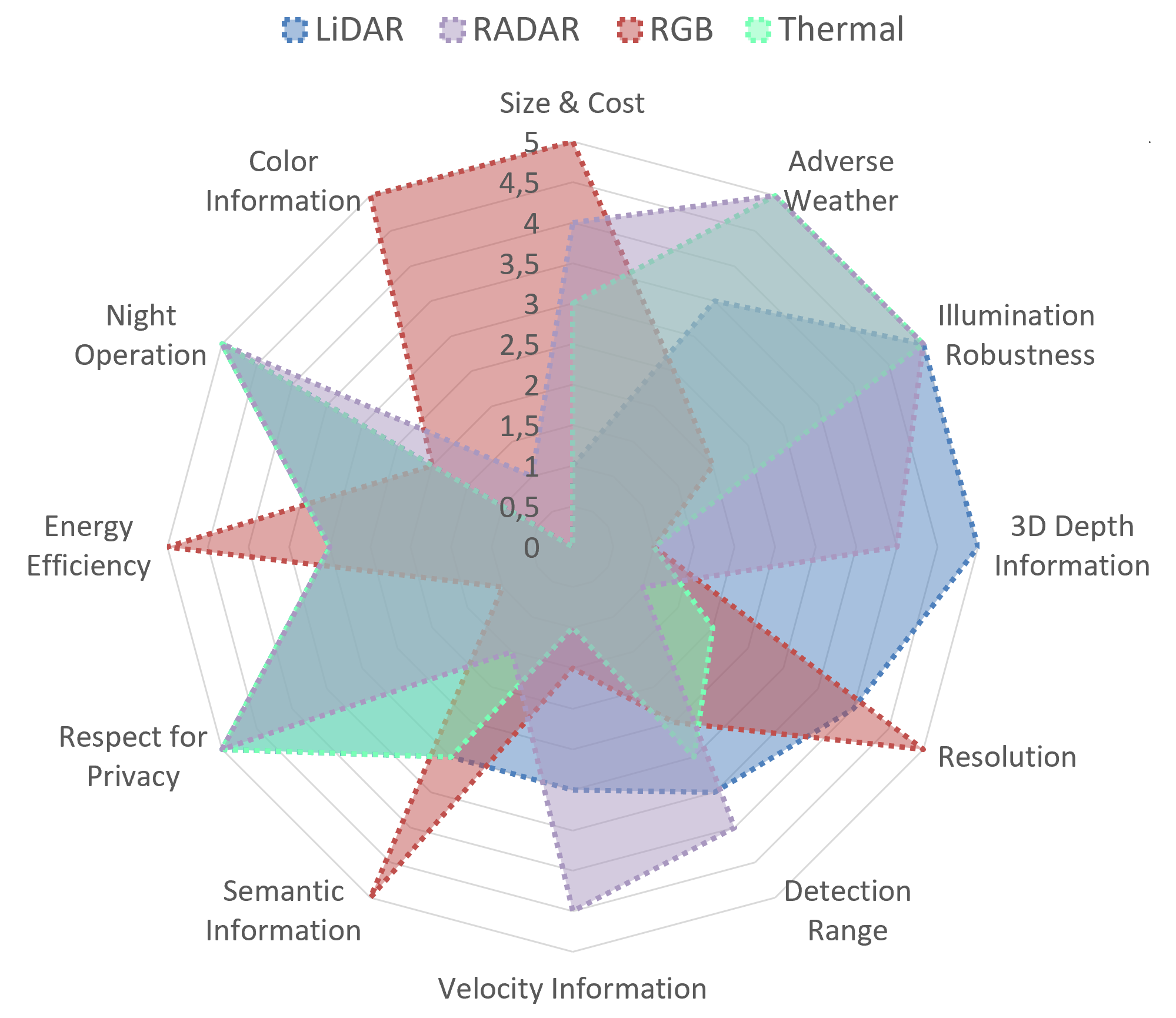}
    \caption{Illustration of the characteristics of LiDAR, RADAR, RGB camera, and thermal sensors. Contextually, LiDAR-RADAR-RGB-Thermal fusion has the highest complementary properties.}
    \label{fig:radar_LRCT}
\end{figure}

There is a large corpus of work on integrating diverse sensor modalities to enhance redundancy and achieve more robust scene perception. This is because each sensor comes with inherent strengths and limitations~\cite{EZFusion2022,sezgin2023safe}, as illustrated in Fig.~\ref{fig:radar_LRCT}.
For instance, RADAR demonstrates strong resilience to severe weather, as its long-wavelength electromagnetic signals can more effectively penetrate atmospheric disturbances~\cite{sezgin2023safe}. 
This robustness enables reliable operation. However, RADAR may also produce spurious detections when reflections from raindrops are misinterpreted as valid returns, particularly at short ranges.
In contrast, LiDAR offers highly accurate 3D spatial measurements under clear conditions, but its performance deteriorates in dense fog or heavy precipitation due to increased light scattering, especially at longer distances.
RGB cameras provide rich visual information, including color and contextual cues, which are essential for semantic scene understanding. However, their reliability declines significantly under harsh weather and poor lighting conditions. 
Thermal cameras, by capturing temperature-based information, remain effective in low-light and nighttime environments, where RGB cameras struggle (see Fig.~\ref{fig:radar_LRCT}).
Fusing these complementary sensing modalities can help leverage their respective strengths while mitigating individual limitations, ultimately improving perception performance in all-weather scenarios~\cite{EZFusion2022}.

To filter out the weather-induced noise (i.e., outliers caused by falling snowflakes or raindrops) in raw sensor readings, contemporary approaches heavily rely on heuristic, statistically driven filtering methods~\cite{kurup2021dsor,huang2023lidsor,park2020fast,rusu20113d,charron2018noising} or advanced data-driven deep learning models~\cite{raisuddin20243d,3DUnOutDet2024,zhao2025tiple,yu2022lisnownet}.
However, these methods exhibit two fundamental limitations. 
First, most approaches require supervision signals, i.e., costly, labor-intensive noise annotations of raw data. 
Second, all these methods are inherently unimodal, operating solely on either 3D LiDAR point clouds~\cite{raisuddin20243d,3DUnOutDet2024,zhao2025tiple,yu2022lisnownet} or 2D RGB images~\cite{Dong_2025_CVPR,DesnowFormer2022,DesnowNet2018}, and consequently under-exploit complementary information available across other sensing modalities.

As of 2026, there is no established consensus on a metric to quantify data quality levels, akin to Technology Readiness Levels (TRLs)~\cite{annex_trl}. However, as part of the ROADVIEW project, we developed a single-value(d) assessment across an arbitrary sized dataset with an arbitrary number of metrics~\cite{Marsh:IV20265}. 

\textbf{Challenge.} Consequently,  weather-induced degradation across sensing modalities poses several challenges: 
\textbf{\textit{(i) how can we learn a generic, sensor-agnostic filtering method that can be applied to diverse sensor readings without requiring costly noise annotations?}} and  
\textbf{\textit{(ii) how can we reliably quantify the quality of both raw and filtered sensor readings?}}
Addressing these challenges is crucial for improving the robustness of downstream perception tasks under adverse weather conditions.

\begin{figure*}[!t]
    \centering
    \includegraphics[width=1\linewidth]{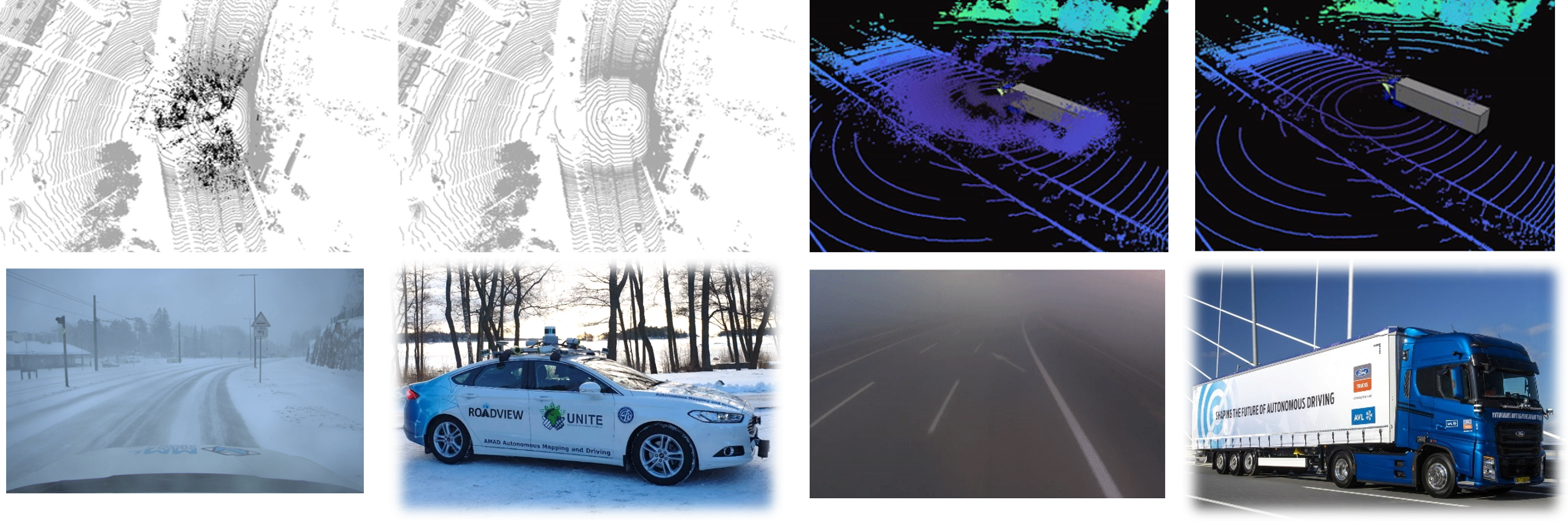}
    \caption{3D point cloud denoising in two ROADVIEW demonstrations: a passenger vehicle operating under snowfall in Finland (left) and a truck driving on a rainy highway in Turkey (right). The top row shows the point clouds before and after denoising. Corresponding RGB images are provided for visualization purposes only. 
    }.
    \label{fig:denoising}
\end{figure*}

\textbf{Recommendation.} Very recently, the ROADVIEW consortium introduced various supervised~\cite{raisuddin20243d} and unsupervised~\cite{Vezeteu2025, 3DUnOutDet2024} methods for 3D point cloud denoising, which can also operate in a multimodal setting~\cite{3D-MMOutDet} by jointly exploiting LiDAR, RGB, and thermal data. Fig.~\ref{fig:denoising} shows sample denoised point clouds from two different ROADVIEW demonstrations.
Our extensive experimental evaluations on multiple datasets, including CADC~\cite{pitropov2021canadian}, WADS~\cite{kurup2021dsor}, and Snowy Scenes~\cite{snowyscenes}, show that early filtering of weather-induced noise in raw sensor readings yields substantial improvements in downstream perception tasks. Specifically, it improves semantic segmentation performance by over $5\%$ mIoU~\cite{raisuddin20243d,3DUnOutDet2024,3D-MMOutDet} and object detection by approximately $2\%$ mAP~\cite{3DUnOutDet2024,3D-MMOutDet}, without requiring any modifications to the original segmentation or detection models.

Building on our findings from ROADVIEW, we encourage the research community to focus on (i) early denoising of multimodal sensor measurements before feeding them to downstream models, (ii) the development of unsupervised filtering methods to avoid costly noise annotations, and (iii)  standardized metrics to quantify the quality of sensor data.

\subsection{Real-time weather-aware multimodal perception}
\label{sec:realtimeperception}
 
Adverse weather phenomena such as snow, rain, fog, and low-visibility conditions affect sensing modalities differently, leading to heterogeneous degradations in downstream perception tasks. 
Increasing redundancy in the perception stack by introducing multiple sensing modalities can significantly improve perception robustness, since different modalities provide complementary information, as illustrated in Fig.~\ref{fig:radar_LRCT}.
Thus, effective perception systems must not only fuse information from multiple sensors but also continuously assess and adapt to the changing environmental context.

A fundamental limitation of many existing multimodal perception frameworks is their lack of scene- and context-aware adaptation. Most approaches rely on static fusion strategies, where the relative importance of each sensor modality is predefined during training and remains fixed during deployment. Thus, these systems cannot appropriately adjust their fusion behavior when individual sensors are degraded by environmental factors. As depicted in Fig.~\ref{fig:radar_LRCT}, RGB-only perception, for instance, may deteriorate under low-visibility conditions, while LiDAR measurements can become heavily contaminated by snowflakes or rain-induced artifacts. Treating all sensor inputs equally under such circumstances can significantly reduce overall perception reliability.

\textbf{Challenge.} Consequently,  multimodal perception involves several challenges: 
\textbf{\textit{(i) how can we learn the strengths and limitations of each sensing modality across diverse environmental conditions?}}   
\textbf{\textit{(ii) how can the perception system dynamically adapt to changing weather conditions?}} and   
\textbf{\textit{(iii) how can we maintain real-time performance when the number of sensing modalities increases?}}
Addressing these challenges is essential for achieving weather-resilient autonomous systems.

\textbf{Recommendation.}
Recent works address these challenges through adaptive fusion mechanisms, including attention-based cross-modal architectures \cite{bai2022transfusion} and availability-aware sensor weighting strategies \cite{paek2026availabilityaware}. While these approaches demonstrate improved resilience to partial sensor degradation, their deployment in safety-critical autonomous systems remains challenging due to the substantial computational overhead associated with real-time inference. 
We therefore recommend prioritizing solutions that balance perception accuracy, adaptability, latency, and energy efficiency to enable real-time, weather-resilient multimodal perception.

Within ROADVIEW, we developed and demonstrated various multimodal perception solutions that incorporate high-level environmental awareness capabilities, including weather-type classification, road-surface condition and slipperiness estimation, and visibility-range assessment. 

\begin{figure}[!b]
    \centering
    \includegraphics[width=1.0\linewidth]{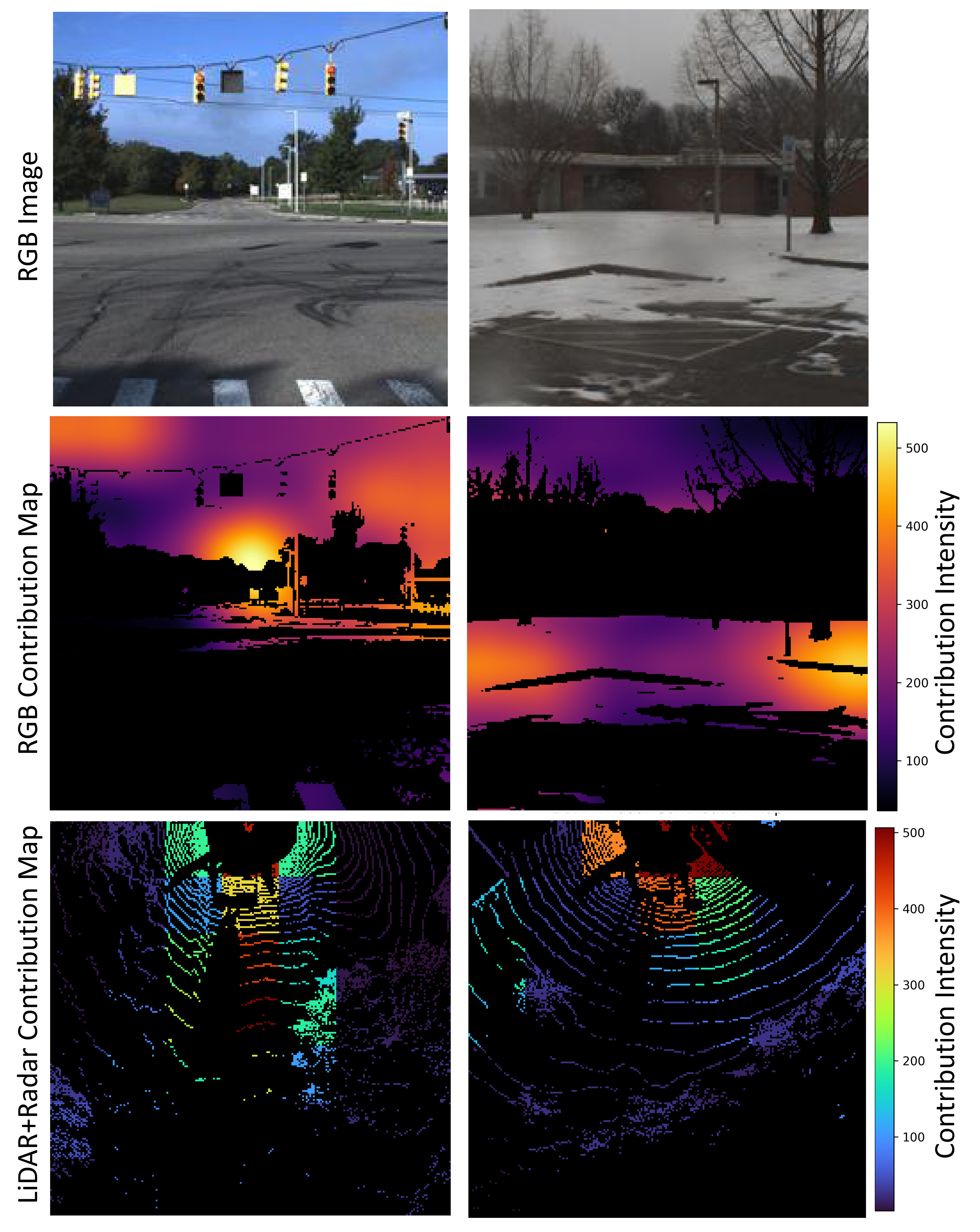}
    \caption{Visualization of weight maps learned in the adaptive gated fusion mechanism in LRC-WeatherNet~\cite{lrcWeatherNet}  for two sample scenes: clear daytime (left) and snowy (right) scenes.    
    }
    \label{fig:lrc_gating}
\end{figure}

For instance, our multi-sensor fusion framework LRC-WeatherNet~\cite{lrcWeatherNet} integrates LiDAR, RADAR, and camera data for real-time weather-type classification.
By combining early fusion through a unified Bird's-Eye-View representation with mid-level gated fusion of modality-specific feature maps, LRC-WeatherNet adapts to the varying reliability of each sensor under changing weather conditions. 
As illustrated in Fig.~\ref{fig:lrc_gating}, due to its adaptive fusion strategy, LRC-WeatherNet relies more heavily on the spatial and reflectivity information from LiDAR and RADAR in visually ambiguous or low-contrast environments, whereas RGB information becomes increasingly dominant in well-lit scenes with clear visual features.

\begin{figure*}[!t]
    \centering
    \includegraphics[width=1.0\linewidth]{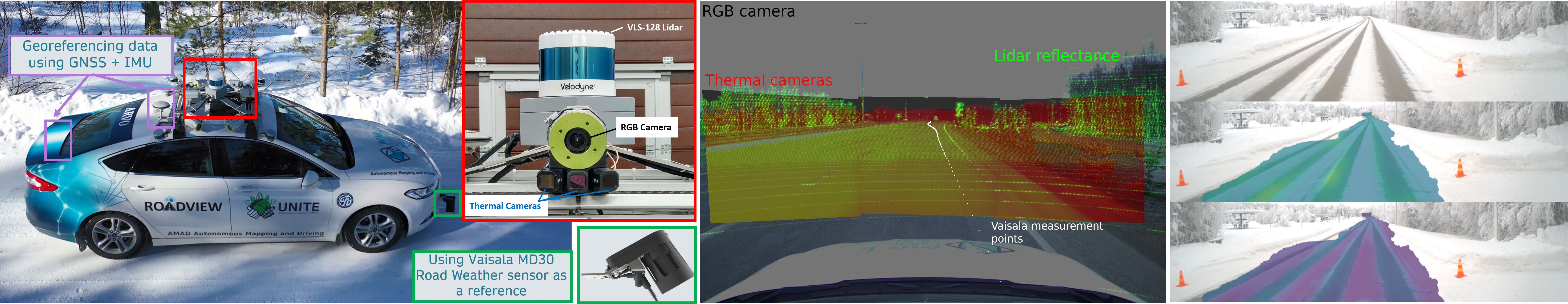}
    \caption{Multimodal uncertainty-aware slipperiness estimation~\cite{maanpaa2024dense}. ROADVIEW vehicle platform equipped with a 360-degree, high-resolution 128-beam LiDAR, a front-facing RGB camera, three thermal cameras, GNSS/IMU, and a Vaisala MD30 sensor (left). All thermal, LiDAR, and Vaisala data are projected to the RGB image space (middle). A sample road surface grip prediction (top: RGB, middle: predicted grip value, bottom: predicted grip uncertainty) (right). The predicted grip value is the mean of the predicted grip probability distribution, and lighter colors indicate more slippery surfaces. The predicted grip uncertainty is the distance between the mean of the grip distribution and the 5th percentile. Likewise, lighter colors represent higher prediction certainty.
    }
    \label{fig:mmslipperiness}
\end{figure*}

The ROADVIEW slipperiness prediction solution~\cite{maanpaa2024dense} (see Fig.~\ref{fig:mmslipperiness}) estimates the road surface grip of snow-covered roads. This grip predictor is based on a convolutional neural network trained using RGB camera images and grip measurements obtained from an optical road-weather sensor (Vaisala MD30). The model predicts a probability distribution of grip values for each image pixel, which is subsequently transformed into a local grip map using both the estimated mean grip value and the 5th percentile of the predicted distribution~\cite{maanpaa2024dense}. The method can successfully distinguish regions with varying levels of road grip and identify areas associated with higher uncertainty, such as tire tracks that may contain thin layers of ice. In many scenarios, the estimated grip remained uniform across the road surface despite visible variations in surface appearance. An example prediction is shown in Fig.~\ref{fig:mmslipperiness}.

Recalling the multimodal and multitask datasets Snowy Scenes~\cite{snowyscenes} and M3 Scenes~\cite{m3scenes} introduced in Section~\ref{sec:sec_dataset}, we encourage the research community to develop unified multitask perception models capable of jointly solving multiple perception tasks, such as 3D object detection and semantic segmentation, within a single forward inference pass. By sharing feature representations across tasks, such models can significantly reduce computational costs, inference latency, and hardware resource requirements compared with executing separate models for each perception task.
%
%
\subsection{Weather-aware map-based positioning}
\label{sec:mapbasedpositioning}



Autonomous vehicle positioning commonly relies on high-end satellite navigation sensors \cite{he2023research} or High-Definition (HD) maps that represent road infrastructure, such as the road itself, road lanes, and traffic signs \cite{javanmardi2018factors,elghazaly2023high}. 
Map-based positioning can provide reliable localization when the satellite navigation-based methods fail due to signal loss in urban canyons, tunnels \cite{jiang2025vehicle}, or forests \cite{kaartinen2015accuracy}, or due to false signals caused by multi-path signals or malicious spoofing of satellite signals \cite{zidan2020gnss}.
Although map-based positioning serves as an alternative to satellite navigation, traditional HD map solutions cannot deliver accurate positioning in challenging conditions, for instance, when road and lane markings are obscured by snow. Alternatively, a map based on LiDAR measurements of the complete road environment can enable accurate positioning even in varying lighting and weather conditions. 

\begin{figure}[!b]
    \vspace{-12pt}
    \centering
    \subfloat[Original KITTI]{%
        \includegraphics[width=0.489\linewidth]{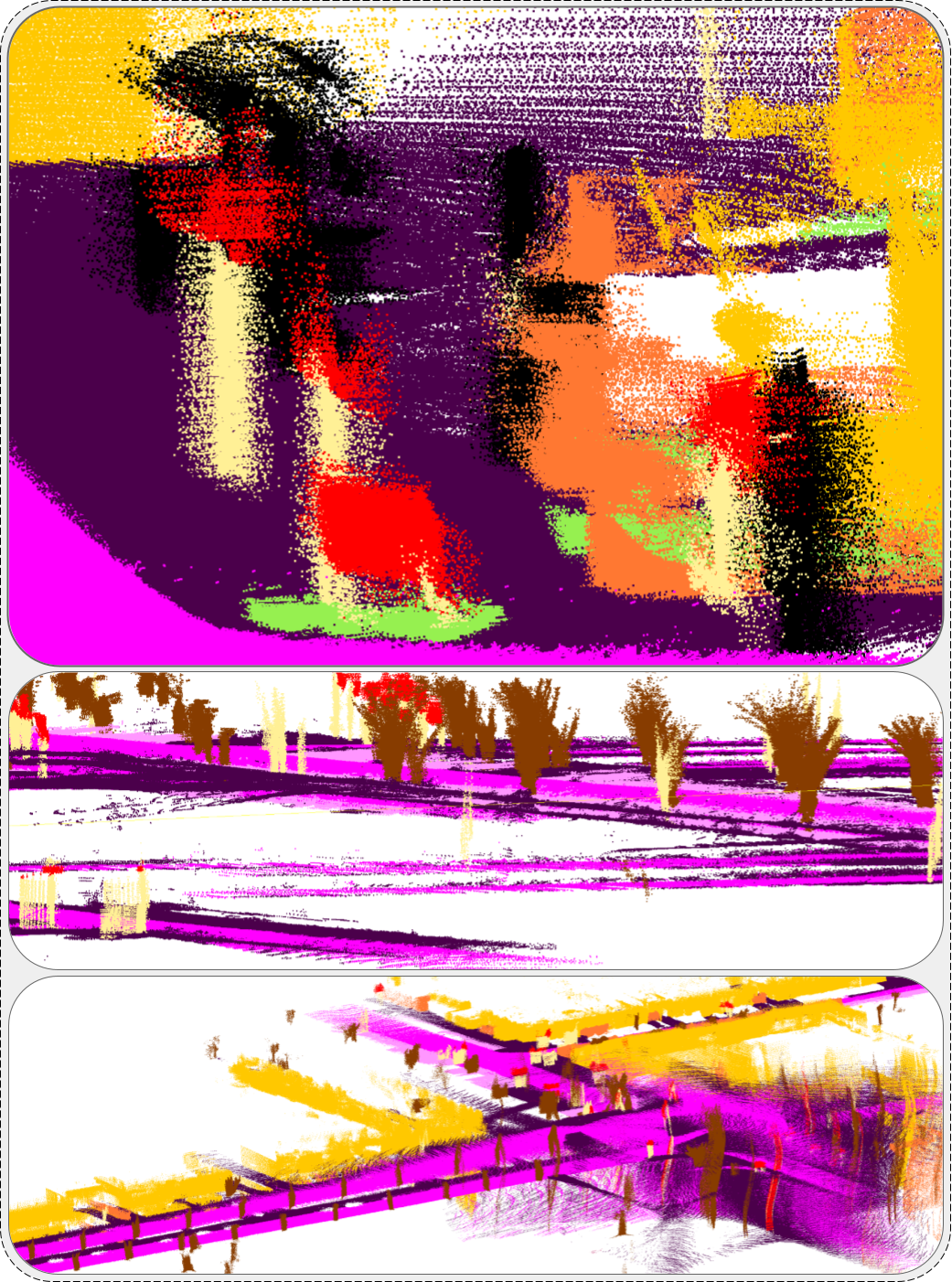}
        \label{fig:abstract_kitti}}
    \subfloat[Revised FGI-KITTI]{%
        \includegraphics[width=0.489\linewidth]{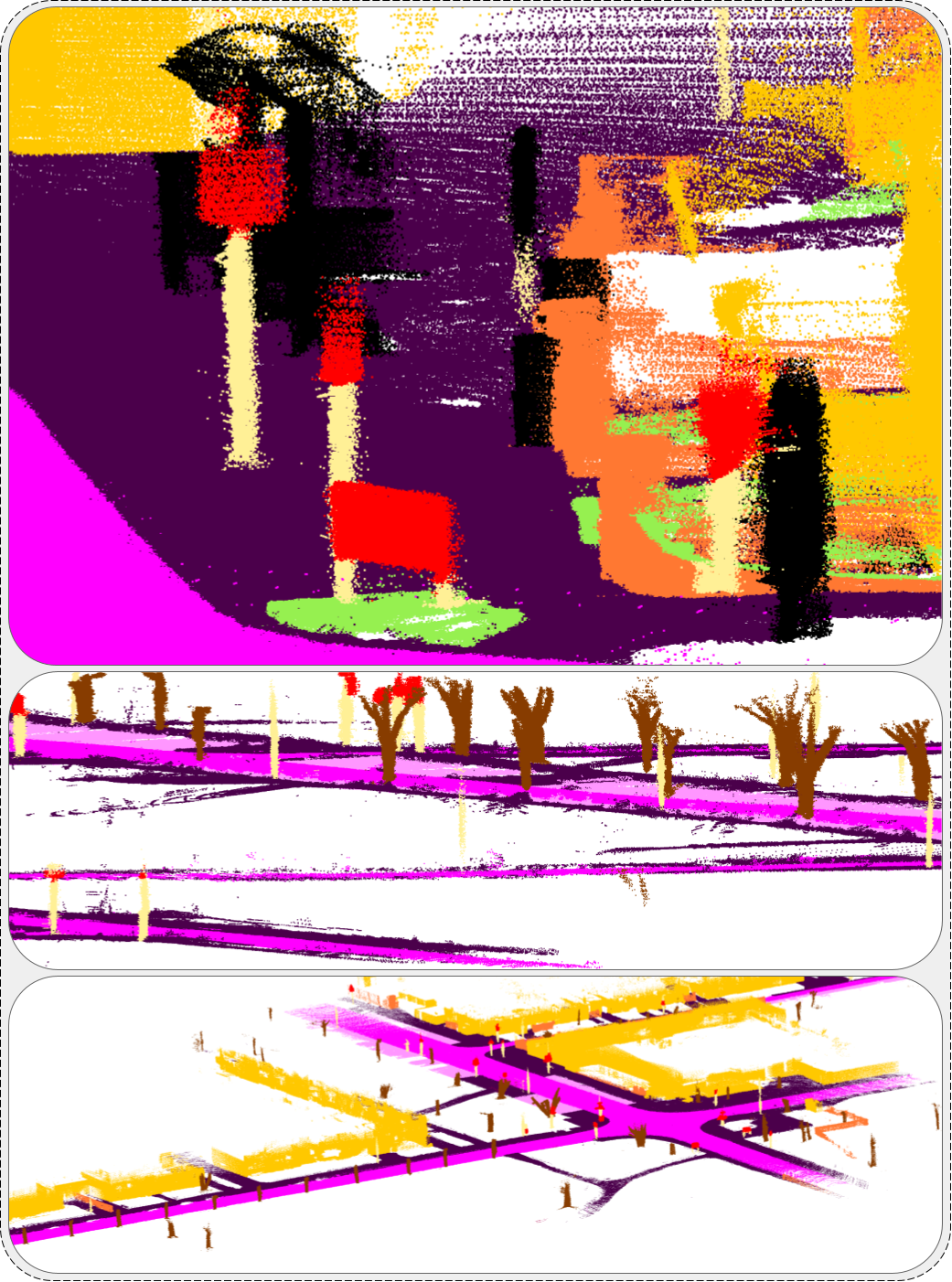}
        \label{fig:abstract_fgi_kitti}}
    \caption{Accumulated semantic LiDAR point clouds located using the GNSS/INS measurements (a) and GLO-SAM-corrected trajectory (b). The corrected reference trajectory results in substantially improved global consistency of the accumulated point cloud.}
    \label{fig:visual_abstract}
    \vspace{-5pt}
\end{figure}

\textbf{Challenge.} Reliable map-based positioning becomes challenging under adverse conditions such as snowfall, rain, and poor visibility, which can increase noise in LiDAR measurements and reduce the amount of observable environmental structure. 
Conventional HD maps based on road and lane markings can also become unreliable when these features are obscured by snow or otherwise changed. 
LiDAR-based HD maps can provide an alternative by representing the complete road environment, but there exist several challenges that must be addressed to achieve accurate and robust positioning: 
\textbf{\textit{(i)  the LiDAR data must be accurately georeferenced}}, as GNSS/INS does not provide a sufficiently accurate reference trajectory for map building, even under good weather and signal conditions, as illustrated in Fig.~\ref{fig:abstract_kitti}.
\textit{\textbf{(ii) The lack of an absolute ground truth }}also makes the evaluation and comparison of reference trajectories difficult.
\textbf{\textit{(iii) In addition, snowfall, rain, and dynamic objects such as vehicles and pedestrians need to be removed}} from the LiDAR data before generating the map from the point cloud.
\textbf{\textit{(iv) Furthermore, the resulting point cloud map should be sufficiently compact}} without compromising positioning accuracy, and the positioning solution must remain robust to seasonal changes, such as snow cover and snowbanks.

\begin{figure}[!t]
    \centering
    \includegraphics[width=0.99\linewidth]{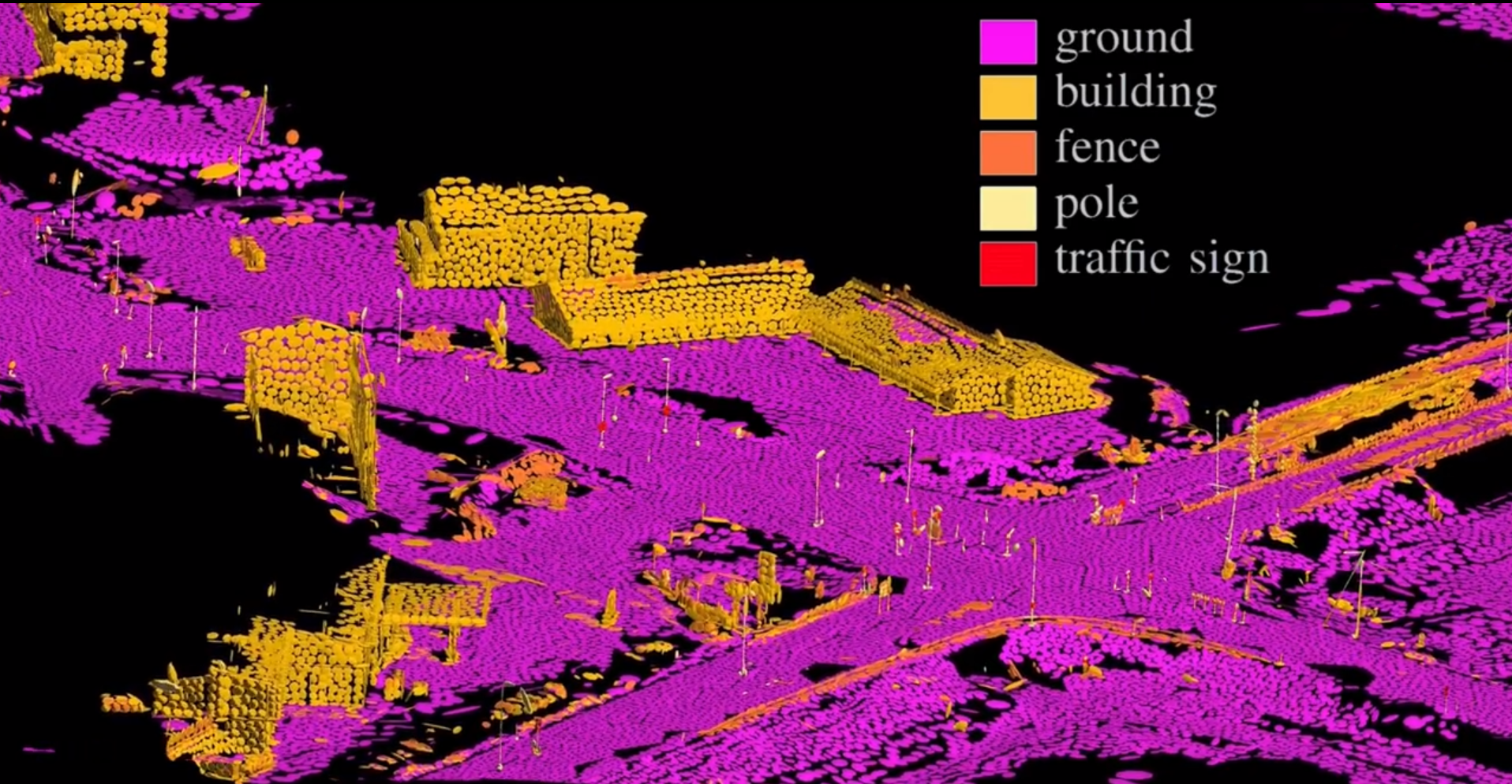}
    \caption{Visualization of the EA-NDT HD Map representation from Muonio centre, northern Finland. The data were recorded with the FGI ARVO autonomous research platform visible in Figs.~\ref{fig:multi_domain} and \ref{fig:denoising}. The entire mapped area of Muonio is described in more detail in \cite{rvD82}.}
    \label{fig:ea_ndt_hd_map}
\end{figure}

\textbf{Recommendation.} Based on our extensive evaluation, we recommend using HD map-based positioning as a complementary solution to GNSS/INS in autonomous driving. 
In ROADVIEW, the consortium demonstrated Environment-Aware Normal Distributions Transform (EA-NDT)~\cite{manninen2022towards, rvD56, rvD57} HD map positioning in three real-world cases~\cite{rvD82, rvD84, rvD85}.
The EA-NDT HD map, visualized in Fig.~\ref{fig:ea_ndt_hd_map}, represents semantically segmented LiDAR data as a data-efficient 3D representation of normal distributions.
LiDAR scans were denoised to mitigate snowfall and rain \cite{raisuddin20243d, Vezeteu2025, 3DUnOutDet2024, 3D-MMOutDet} and Cylinder3D~\cite{zhou2020cylinder3d}, trained with ROADVIEW Snowy Scenes \cite{snowyscenes}, was used for semantic segmentation.

EA-NDT positioning was demonstrated using a truck (Fig.~\ref{fig:denoising}) on a highway in Istanbul, Turkey~\cite{rvD84}, a passenger vehicle operating in snowy conditions in Muonio, northern Finland~\cite{rvD82}, and a van in Roding, southeast Germany, under fog and rain in AVL’s weather hall~\cite{rvD85}. 
In all demonstrations, EA-NDT provided better orientation accuracy than GNSS/INS and nearly comparable position accuracy, as reported in~\cite{rvD88}. 

\begin{figure}[!b]
    \centering
    \subfloat[]{%
        \includegraphics[width=0.98\linewidth]{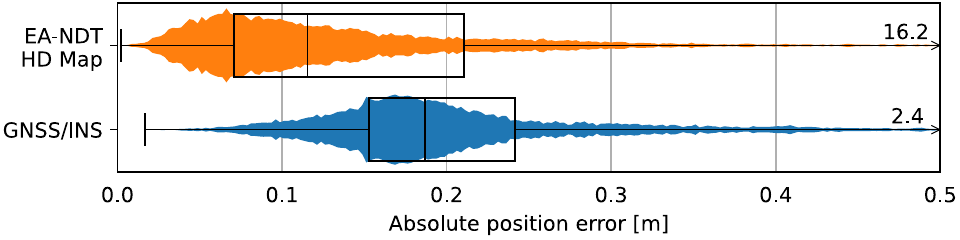}
        \label{fig:eandt_bp_abs_pos_err}}
        
    \subfloat[]{%
        \includegraphics[width=0.98\linewidth]{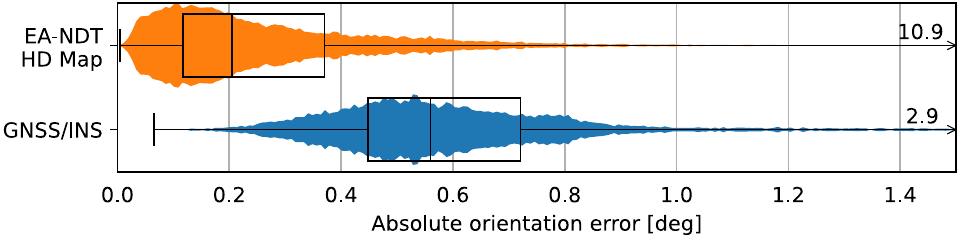}
        \label{fig:eandt_bp_abs_rot_err}}
    \caption{A box-and-whisker plot of absolute position (a) and absolute orientation (b) errors along an 11-km-long trajectory in Muonio, northern Finland.}
    \label{fig:eandt_bp}
\end{figure}

\begin{figure}[!t]
    \centering
    \includegraphics[width=0.99\linewidth]{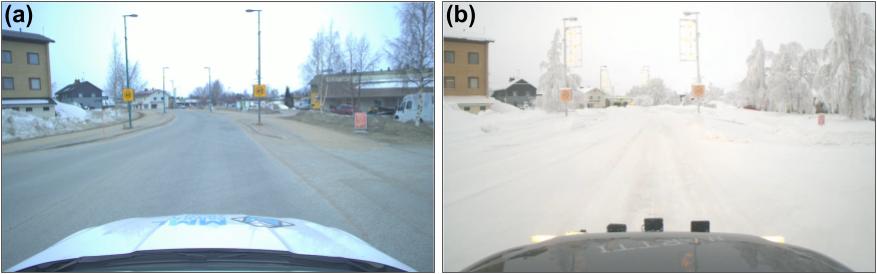}
    \caption{The visual difference of the environment demonstrated with a front camera view of the same scene in the centre of Muonio in (a) during the EA-NDT map data collection in April 2025 with the FGI ARVO vehicle, and (b) during the EA-NDT positioning demonstration in January 2026 with the VTT Martti vehicle.}
    \label{fig:muonio_data_cameras}
\end{figure}

In GNSS-denied areas, EA-NDT provided an independent positioning solution with higher accuracy than GNSS/INS.
For the 11.0 km Muonio trajectory, Fig.~\ref{fig:eandt_bp} shows the distribution of absolute positioning and orientation errors. 
The results indicate that EA-NDT generally provided higher positioning accuracy than GNSS/INS, despite some larger outliers. 
These outliers degraded the performance at a few locations due to changes between the mapped and observed environments, as illustrated in Fig.~\ref{fig:muonio_data_cameras}. 
Addressing such environmental changes, therefore, remains an important challenge for improving the robustness and overall accuracy of HD map-based positioning.

To achieve the positioning accuracy demonstrated above, the LiDAR data used to build the EA-NDT HD map must be accurately georeferenced, i.e., aligned with a global coordinate system. We therefore strongly encourage the combination of satellite navigation with accurate LiDAR odometry and loop-closure constraints for map georeferencing. The ROADVIEW consortium recently introduced Global LiDAR Odometry Smoothing and Mapping (GLO-SAM)~\cite{manninen2026toward}, a method for accurate georeferencing of LiDAR data that also provides point-cloud geometry-based metrics for comparing reference trajectories. GLO-SAM uses a Graph-SLAM approach to jointly optimize GNSS, LiDAR odometry, and loop-closure constraints to build a coherent, georeferenced point cloud map. This approach was used to georeference the LiDAR data in the ROADVIEW demonstrations with EA-NDT HD map positioning.


\begin{figure}[!b]
    \centering
    \includegraphics[width=0.99\linewidth]{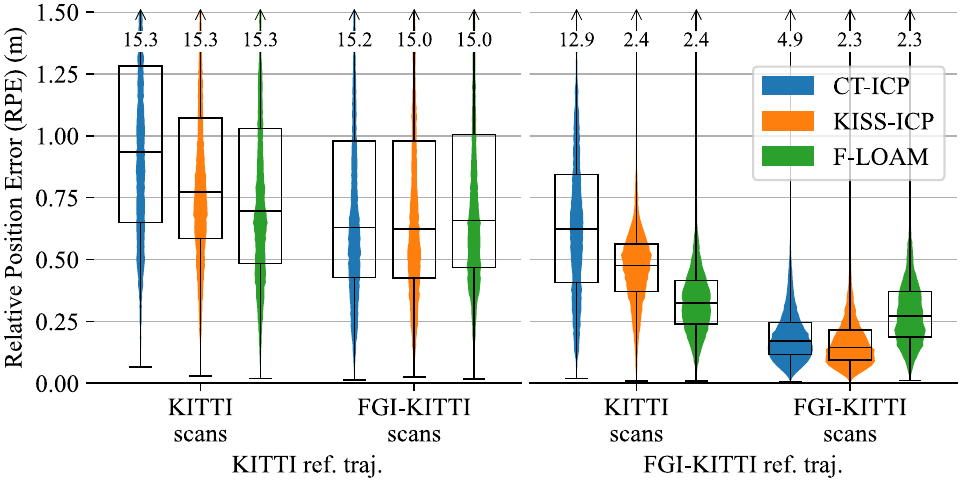}
    \caption{A box-and-whisker plot of RPE for the odometry benchmark using CT-ICP, F-LOAM, and KISS-ICP. The methods are evaluated using the original KITTI and revised FGI-KITTI scans against both the original KITTI and revised FGI-KITTI reference trajectories. The results demonstrate the effect of reference-trajectory accuracy on LiDAR odometry benchmarking.}
    \label{fig:odometry_rpe_box}
\end{figure}

The importance of accurate georeferencing extends beyond HD map generation. 
For instance, GLO-SAM was further applied to the KITTI odometry dataset (captured in clear weather) to improve the georeferencing and consistency of the reference trajectories. In parallel, the LiDAR scans were geometrically corrected to address inconsistencies in the original KITTI LiDAR data. The resulting dataset, referred to as FGI-KITTI, therefore provides both corrected LiDAR scans and an improved reference trajectory. Applying GLO-SAM substantially improved the global consistency of the accumulated point cloud, as illustrated in Fig.~\ref{fig:visual_abstract}. The effects of these corrections are also reflected in the odometry evaluation, where the results obtained with different combinations of LiDAR scans and reference trajectories demonstrate that both the LiDAR data and reference trajectory can influence benchmark results (Fig.~\ref{fig:odometry_rpe_box}). The resulting improvements in point cloud geometry and odometry evaluation support the use of FGI-KITTI as a more consistent basis for benchmarking LiDAR odometry and SLAM algorithms.

%
%
\subsection{Collective perception via V2X communication}
\label{sec:collectiveperception}

Collective perception enables vehicles and road infrastructure to extend their
awareness beyond the sensing range of individual onboard sensors by sharing
perception output through Vehicle-to-Everything (V2X) communication~\cite{v2x}.
Unlike already deployed V2X services that mostly exchange vehicle status
information, Collective Perception Messages (CPMs)~\cite{ETSI_TS_103_324} allow
the dissemination of detected objects, free-space information, and perception
metadata generated by onboard or infrastructure-based sensors.

\begin{figure}[!b]
    \centering
    \includegraphics[width=1.0\linewidth]{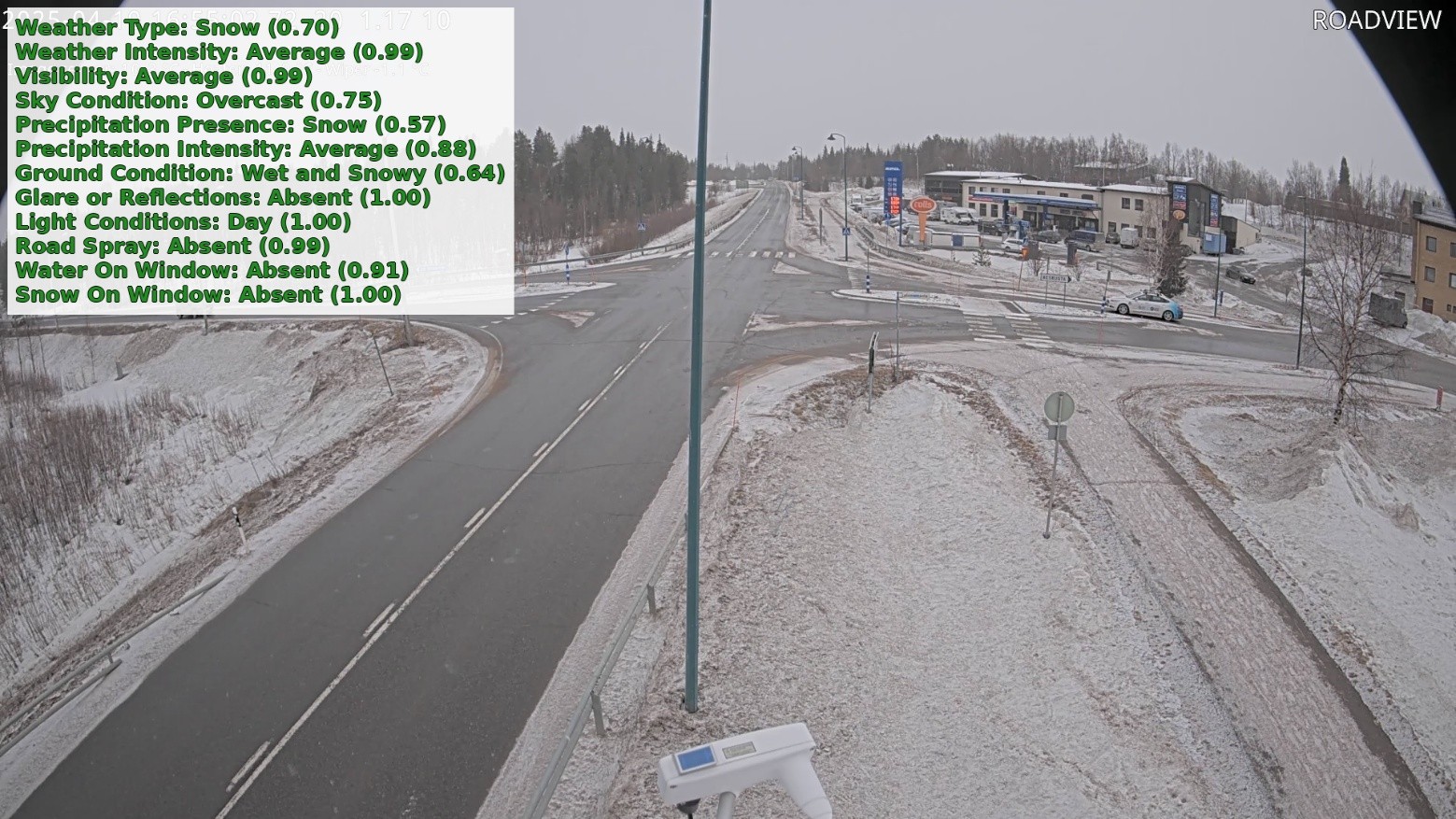}
    \caption{Camera-based roadside weather detection: weather type detected as snow, ground conditions as wet and snowy.}
    \label{fig:roadside_weather}
\end{figure}

Beyond simply transmitting traffic conditions to vehicles via V2X
communications, it is crucial to anticipate the meteorological conditions that
vehicles will encounter ahead. Acquiring this real-time information requires
robust collection methodologies. Currently, conventional approaches rely on the
deployment of road weather stations. However, these infrastructures are
expensive and are not regularly spaced along road networks. To address these
limitations, we propose, within the framework of the ROADVIEW project, the
development of innovative methods to detect and characterize weather conditions
directly from roadside cameras
\cite{ouattara2026stylebasedneuralarchitecturesrealtime,ouattara2026stylisticstormststormperceiving,ouattara2026heuristicstyletransferrealtime,Ouattara2026c,Ouattara2026d,Ouattara2025},
which have a much denser spatial distribution along road networks.
Fig.~\ref{fig:roadside_weather} illustrates an example of a roadside
camera-based weather detection system. Furthermore, the developed system
integrates the capability to leverage vehicle-mounted cameras. This collaborative approach enables a joint and shared perception of the road
environment between the infrastructure and connected vehicles, thereby
facilitating the dynamic dissemination of this safety information through
future standardized V2X messages.

\begin{figure}[!b]
    \centering
    \includegraphics[width=1.0\linewidth]{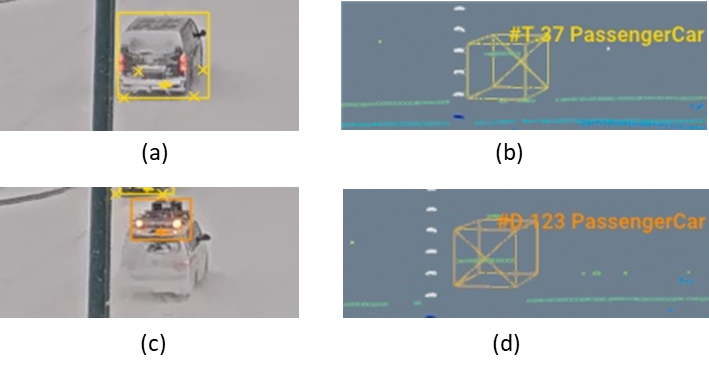}
    \caption{Multimodal sensing in snowy conditions (a) Camera-based detection system correctly detecting the black car; (b) LiDAR-based detection system correctly detecting the black car; (c) Camera-based detection system partially detecting the white car due to snow spray; (d) LiDAR-based detection system correctly detecting the white car.}
    \label{fig:multi_modal_sensing}
\end{figure}

Within ROADVIEW, a multimodal roadside perception system, combining cameras and LiDAR sensors, was deployed in Northern Finland to investigate the impact of
harsh winter conditions on collective perception and to evaluate how
weather-dependent perception quality should be represented in
CPMs~\cite{rvD54}. The experiments highlighted the complementarity of
multimodal sensing. Under snowy conditions, camera-based perception may suffer
from reduced visibility or low contrast, whereas LiDAR can remain capable of
detecting objects. Fig.~\ref{fig:multi_modal_sensing} shows an example where
the camera fails to detect a white vehicle obscured by snow spray,
while the LiDAR successfully identifies the same target. Such complementary behavior motivates the use of multimodal roadside sensing and illustrates why
confidence indicators should accompany perception outputs transmitted through
CPMs. When sharing sensor information using V2X, it is important to include
metadata about the sensor's performance and the environmental conditions at the
time of data capture. This allows receiving vehicles to assess the reliability
of the shared data and make informed decisions about its use in their own
perception systems.

\begin{figure*}[!t]
    \centering
    \includegraphics[width=1.0\linewidth]{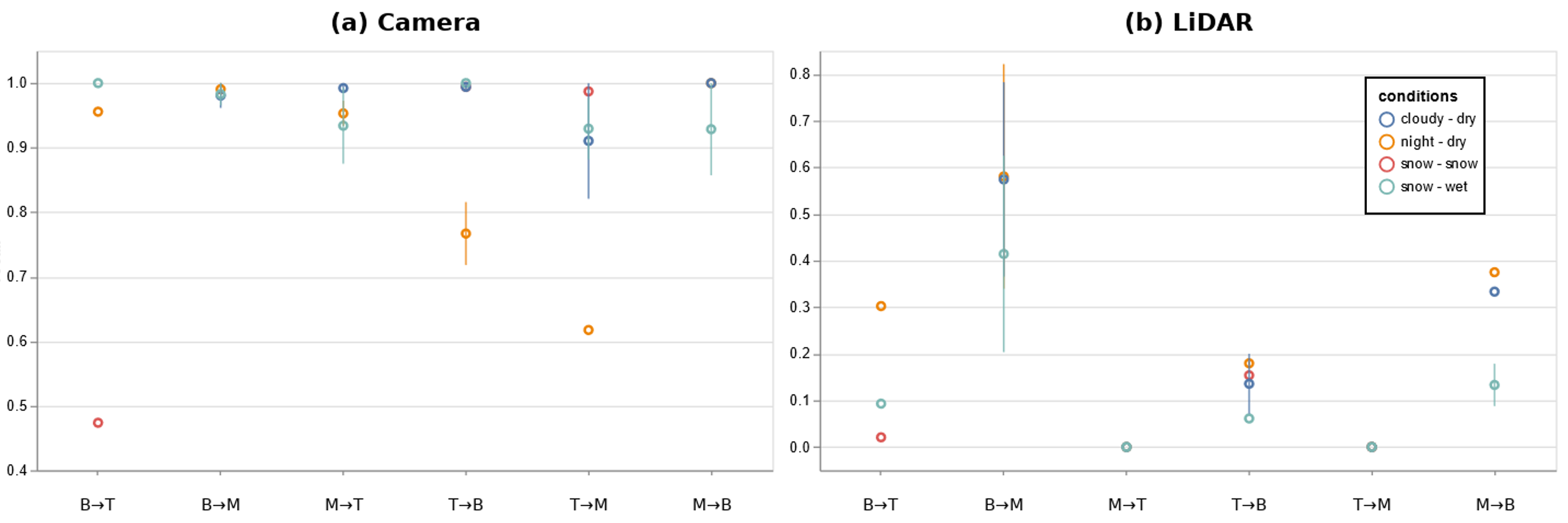}
    \caption{
        Recall of roadside camera-based (left) and LiDAR-based (right)
        vehicle detection systems measured under different weather conditions
        and vehicle trajectories. The x-axis shows the recall score, while the y-axis represents the vehicle trajectory (path). Results show that perception performance is dependent on both weather
        conditions and vehicle trajectories, highlighting the need for
        confidence indicators and quality metadata in collective perception
        messages. Path labels are abbreviated as B=bottom, M=middle, and T=top.  
    }
    \label{fig:recall_camera_lidar}
\end{figure*}

To quantify these effects, ROADVIEW conducted measurements in harsh winter
conditions using a probe vehicle providing ground truth. Recall metrics were
evaluated for both camera- and LiDAR-based perception systems under different
weather conditions, as shown in Fig.~\ref{fig:recall_camera_lidar}. The results
demonstrate that sensor performance varies significantly with environmental
conditions and vehicle trajectories. LiDAR recall values are lower than camera
recall values in the reported experiments, which were conducted using a VLP‑16
LiDAR sensor. These observations confirm that perception quality is weather-
and context-dependent. Thus, it should be represented through confidence
indicators and uncertainty metadata in CPMs.

\textbf{Challenge.}
ROADVIEW identified three major challenges for the deployment of weather-aware
collective perception.

\textbf{\textit{(i) Weather-dependent perception uncertainty:}}
The accuracy of roadside perception systems changes with weather and lighting
conditions. Without explicit quality descriptors and environmental context,
receiving vehicles cannot reliably assess the trustworthiness of CPM content.

\textbf{\textit{(ii) Spatial-temporal consistency and calibration:}}
Effective fusion between onboard perception and shared CPM information requires
accurate sensor calibration, object localization, and time synchronization.
Systematic positioning errors or temporal misalignment reduce fusion accuracy
and may lead to inconsistent world models.

\textbf{\textit{(iii) Scalability and interoperability:}}
Large-scale deployment of CPM-capable stations introduces challenges related to
radio-channel congestion, information relevance, and interoperable
implementation of CPM standards. Consistent interpretation of quality fields, coordinate systems, and confidence indicators across vendors remains an open issue.

\textbf{Recommendation.}
Based on ROADVIEW experimentation, we recommend the following directions for
future deployments and standardization efforts.

\textit{(i) Integrate weather and perception-quality information into V2X messages:}
Roadside weather observations and perception reliability indicators should be
shared together with the detected objects. Weather-dependent confidence measures
would allow receiving vehicles to adapt the weighting of shared information
during sensor fusion.

Roadside weather observations and perception reliability indicators should be
disseminated together with detected objects. ROADVIEW demonstrated that
roadside and vehicle-mounted cameras can provide cost-effective weather
monitoring with much higher spatial coverage than conventional weather
stations. Such information can complement collective perception services and
allow receiving vehicles to adapt the weighting of shared information during
sensor fusion.

\textit{(ii) Maintain accurate calibration and uncertainty characterization:}
Roadside sensing systems should rely on robust calibration procedures,
continuous monitoring of systematic errors~\cite{cancouet_mitigating_2026}, and clear uncertainty
characterization to support reliable fusion with onboard perception.

\textit{(iii) Promote value-of-information-driven communication and CPM
    profiling:}
Transmission strategies should prioritize information that improves situational
awareness while limiting channel load. Continued ETSI standardization and CPM
profiling activities are needed to guarantee interoperability between vehicles,
roadside infrastructure, and suppliers.

Collective perception can significantly improve situational awareness and
safety, particularly in harsh weather conditions where sensors are subject to
degradation. ROADVIEW demonstrated that weather-aware roadside perception systems can provide valuable information not only about detected objects but also about the environmental conditions that affect perception quality. Future large-scale deployment will require robust uncertainty representation within CPMs, accurate calibration, scalable communication strategies, and continued standardization to enable reliable cross-vendor interoperability.
%
%
\subsection{Weather-aware decision making}
\label{sec:decisionmaking}

\begin{figure*}[!b]
    \centering
    \includegraphics[width=1\linewidth]{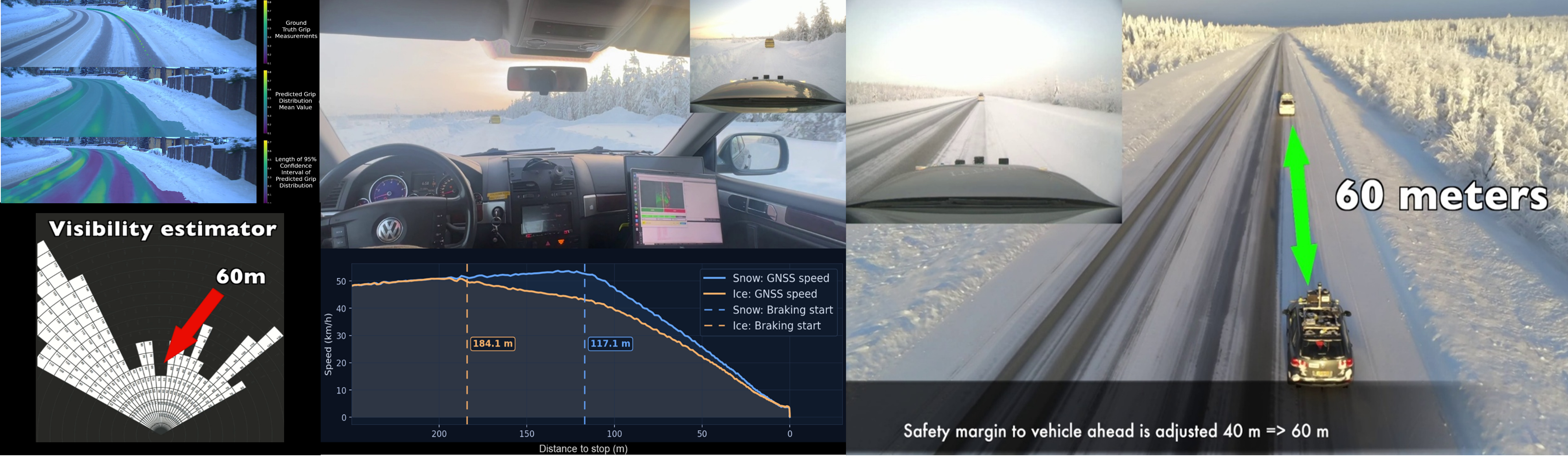}
    \caption{ROADVIEW look-ahead road-condition estimation for weather-aware decision making. The grip predictor~(top-left) estimates local road-surface grip and uncertainty from forward-looking sensor data, while the direction-aware visibility estimator~(bottom-left) assesses the operational perception range under reduced visibility~\cite{rvD82}. 
    In the ROADVIEW demonstration (top-middle), projected road-friction estimates were converted into braking-distance and safety-margin constraints. The vehicle initiated braking earlier on icy surfaces than on snow-covered surfaces (bottom-middle) and increased the following safety margin when reduced friction was estimated (right)~\cite{rvD82}. Click \href{https://www.youtube.com/watch?v=4AKmG4ym_Bc&list=PLPRufXlEiQvriCsh3qalxvxEkPkOmqAES&index=2}{here} to see the demonstration video.
    }
    \label{fig:controller}
\end{figure*}

Automated driving systems must not only perceive their surroundings under adverse weather but also translate degraded environmental conditions into safe driving behavior. In winter conditions, this is particularly challenging because visibility and tyre–road friction may deteriorate simultaneously: heavy snowfall or rain, fog, and sensor contamination reduce the operational range of perception, while ice, snow, slush, and wet surfaces increase braking distance and reduce vehicle controllability. Prior research has addressed these issues from different perspectives. Friction-aware vehicle control has been studied in autonomous vehicle-following and ACC/AEB/ABS functions, where estimated tyre–road friction is used to adapt spacing and braking under low-friction conditions \cite{santini2021board}. Prior work has studied fog detection, LiDAR-based visibility estimation, sensor degradation, and fog-adaptive control, including laboratory visibility estimation \cite{miclea2020laser} and point-cloud-based inference of fog visibility and detection probability \cite{montalban2023bayesian}. However, results obtained in climatic chambers or with predefined targets may not fully capture the variability of real outdoor winter conditions \cite{montalban2023bayesian,kutila2018automotive}. In addition, methods that rely on detailed optical signal behavior may be difficult to generalize across production LiDARs, since sensor-specific processing can affect the available point cloud and intensity data.

In ROADVIEW, these issues were addressed by developing a weather-aware decision maker with velocity control (see Fig.~\ref{fig:controller}). ROADVIEW studies demonstrated that forward-looking RGB, thermal, and LiDAR-reflectance data can support spatially resolved estimation of road-surface grip, using optical road-weather sensor measurements as reference data~\cite{maanpaa2024dense}. In this work, uncertainty was treated as an essential part of grip prediction, since safety-critical control should rely on conservative estimates of available grip rather than only on the predicted mean value~\cite{maanpaa2025road}. In the same context, an in-vehicle visibility estimator was also developed and demonstrated to assess the operational range of the environmental perception system using LiDAR point cloud data under fog, rain, and snow conditions~\cite {rvD82}. The direction-aware visibility estimate was then used to determine whether automated driving could continue or whether a Minimal Risk Maneuver (MRM) was required~\cite{rvD62,rvD63,rvD64}.

\textbf{Challenge.} The main challenge is \textbf{\textit{how to use uncertain and spatially variable estimates of grip and visibility ahead of the vehicle in real-time automated driving decision making}}. Grip and visibility are not direct control commands; they must be converted into actionable constraints such as braking distance, stopping distance, velocity limits, safety margins, and ODD-compliance checks. ROADVIEW showed that braking distance estimation provides a useful, common safety variable that links road-friction estimation, visibility requirements, velocity limitation, and MRM triggering~\cite{rvD62,rvD63,rvD64}. However, both grip and visibility estimates remain uncertain under real-world variable conditions. Therefore, safe control cannot rely only on nominal or mean estimates; uncertainty must be propagated explicitly, and conservative lower-bound reasoning is required when deriving speed limits, safety margins, and stopping distances ~\cite{rvD63,rvD64,maanpaa2025road}. 
%
%
The challenge is not only to estimate road and weather conditions, but also to determine how much the Automated Driving System (ADS) should trust these estimates and how they should affect vehicle behaviour. Extreme weather conditions alter tire-road friction, making accurate vehicle-dynamics prediction difficult.
Robust automated driving therefore requires adaptive vehicle and tire models that can handle changing road conditions and the uncertainties associated with them in real time.

ROADVIEW addressed this challenge by implementing a weather-conditional velocity controller that adapts vehicle speed, acceleration, and braking based on visibility estimates, road-friction estimates, road inclination, and braking-distance estimates~\cite {rvD63}. The evaluation covered cases in which friction reduces safe speed, friction uncertainty is handled using a lower bound, friction affects full-stop and lane-change behavior, and estimated friction changes the safety margin for other road users~\cite{rvD64}.

In one of the ROADVIEW demonstrations~\cite{rvD82}, reduced visibility was generated by a lead vehicle driving at 80 km/h and lifting loose snow from the road surface and creating a dense snow plume; the following automated vehicle detected reduced forward visibility and triggered an MRM when the estimated visibility remained below the required threshold for longer than the defined grace period (see Fig.~\ref{fig:controller}). In the same demonstration, the automated vehicle used projected road-friction estimates to determine safe braking before a stationary obstacle; on icy surfaces, braking started earlier than on snow-covered surfaces, resulting in a more gradual speed reduction. The demonstration also showed safety-distance adaptation, where the following distance increased under snow/reduced-friction conditions compared with dry asphalt~\cite{rvD82}. These results indicate that even imperfect look-ahead estimates are valuable: they allow anticipatory adaptation, whereas without such information, the ADS can only react after the vehicle has already entered the degraded condition.

\textbf{Recommendation.} Future weather-aware ADSs should explicitly propagate uncertainty from grip and visibility estimation into decision-making and control. Conservative estimates should be used for safety-critical computations, especially when deriving stopping distance, speed limits, and following distance~\cite{rvD63,maanpaa2025road}. At the same time, both grip prediction and visibility estimation ahead of the vehicle in real-time require further development before wide deployment. Grip prediction should be improved by using larger, more diverse datasets, especially for rare but safety-critical surface states such as black ice and transitions among snow, slush, water, and ice. Visibility estimation should be further developed towards robust, direction-aware operational visibility assessment using production-relevant automotive sensors, rather than relying only on controlled fog-chamber measurements or assumptions that are difficult to transfer to commercial LiDAR outputs \cite{miclea2020laser}–\cite{kutila2018automotive}. We further recommend converting weather and road-condition estimates into physically meaningful constraints, particularly braking and stopping distance, so that perception, navigation, velocity control, and MRM logic share a consistent safety abstraction~\cite{rvD62,rvD63,rvD64}. Finally, onboard estimation should be complemented with V2X-based road-weather awareness, allowing infrastructure and other vehicles to extend the ADS’s knowledge of upcoming conditions beyond the ego vehicle’s immediate sensing range~\cite{rvD64,rvD82}. 

%
%
\subsection{Sim-to-real gap in noise modelling}
\label{sec:noisemodeling}
The closeness of a simulation output is often compared to reality to determine the likelihood, \textit{i.e., the sim-to-real gap}, which indirectly informs how much you can trust the simulation. This gap is often used to assess the output of simulated perception sensor data, which are affected by a myriad of factors, from sensor intrinsics and material properties to the complex noise factors affecting the environment. Furthermore, the process of generating the sensor data needs to be validated, enabling confidence in using simulation to develop, test, and validate systems.

In the ROADVIEW project, we have sought to reduce this gap through the development and improvement of digital models, model properties, and sensor models. Photogrammetry of the CARISSMA and CEREMA weather generation sites was conducted to produce centimeter-accurate, color-accurate digital twins for sensor data generation. The LiDAR and ROADVIEW 4D RADAR sensor models were created and improved based on physics principles, interactions with material properties, and complex scanning patterns to generate realistic sensor data. Furthermore, to achieve the goal of the ROADVIEW project, physics-based and data-driven weather sensor noise models were developed and validated against real data to augment sensor measurements in harsh, adverse weather. These models were further refined using noise model compounding to increase the realism of the sensor data.

\textbf{Challenge.}
There are increasing efforts in the community to reduce the sim-to-real gap for sensor data through improvements to sensors and sensor noise models. However, there are still limitations in assessing the sim-to-real gap due to various factors.

\textit{\textbf{(i) Importance and controllability of real data for validation.} }The evaluation of generated data requires a comparison with real-world data. This requires datasets that produce the evaluated noise at varying intensities (\textit{e.g.}, with gradually increasing snowfall intensity) for generalisability. Ideally, these frames should be `paired frames', where identical scenes are captured with and without the noise, with exactly the same scene content. Even recent datasets for weather, such as the ACDC~\cite{acdc}, which provided `paired frames' for weather in the real world, lacked exact positional alignment, and there were differences in the static and dynamic elements (\textit{e.g.} different moving vehicles in the scene, construction work, \textit{etc}).

\textit{\textbf{(ii) Lack of standardised metrics.} }Various prior works use a combination of visual comparisons, common image quality assessment metrics, and machine-learning perception performance. However, there is no consensus on the validation methodology or on exactly how to quantify the sim-to-real gap.

\textit{\textbf{(iii) Purpose-driven fidelity.}} As increasingly complex models and systems are developed to reduce the sim-to-real gap, we have to ensure that any development is impactful. Complexity may inevitably reduce the model's usability, as it can introduce limitations such as long processing times and increased hardware requirements. Generating extremely realistic data is not inherently beneficial if the performance of the tested system is not sensitive to such fidelity. 

\textbf{Recommendation.}
Based on the ROADVIEW investigations and outcomes from WP3 and WP7, we recommend the following.

\textbf{(i) Controlled paired datasets.} The ROADVIEW project captured the first weather-paired `REHEARSE' dataset~\cite{REHEARSE2024} for varying rain and fog intensities with a validated distribution of weather drop sizes. This was created with a controlled static setup to ensure alignment and scene content. The REHEARSE dataset was used to validate the ROADVIEW weather noise models. However, there are still challenges with the data capture setup, such as varying sky-lighting conditions, no dynamic movement, and assessment of only a single noise source. Further paired datasets need to be captured to enable validation of different noise levels. However, different properties of the noise should be captured and, where possible, quantified. It was observed in the models for both camera and LiDAR that the drop-size distribution of the weather affects the data more than the weather intensity.

\textbf{(ii) Library of metrics.} To tackle the lack of standardised metrics, ROADVIEW compiled and developed a list of image quality assessment metrics (for camera data)~\cite{gummadi2023correlating} and point cloud comparison metrics (for LiDAR and 4D RADAR) for validation~\cite{shahbeigi2024novel}. This evaluation provides a way to validate different features of the sensor data against real-world data. Further work is needed to assess the generated sensor data itself and the sim-to-real gap of the digital twins.

\textbf{(iii) Downstream tasks.} To ensure that the noise models can be run efficiently in real-time for X-i-L testing, they were adapted with pre-processing and simplifications, sacrificing some accuracy of the model. The simplified sensor data were assessed using vehicular motion data and the perception performance of the ROADVIEW system. 
The direct downstream task is generally a perception system for an automated vehicle.
However, the rapid development of AI networks makes it impossible to choose a representative network for evaluation. Instead, it is crucial to understand how features related to noisy data artifacts affect performance and to focus the evaluation on aspects that influence system behavior.

\subsection{XiL testing and validation}
\label{sec:xil_testing}



\begin{figure}[!b]
    \centering
    \includegraphics[width=1\linewidth]{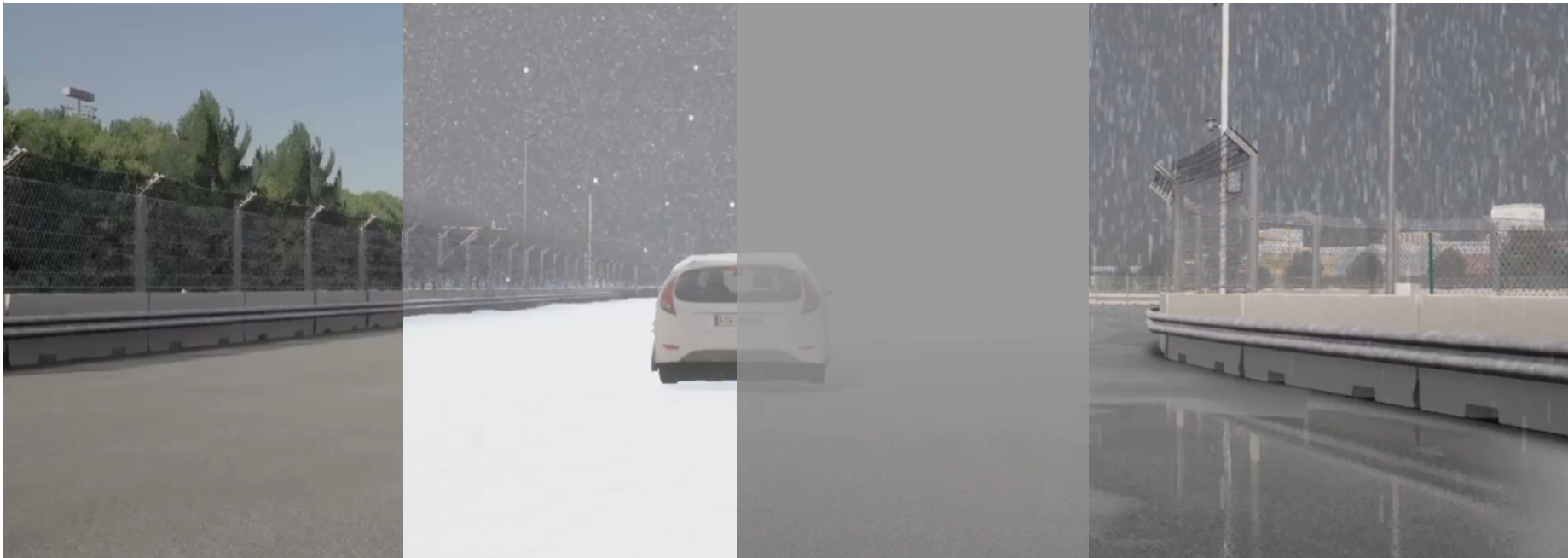}
    \caption{XiL testing of the same scene under different weather conditions: (from left to right) sunny, snowy, foggy, and rainy.}
    \label{fig:XILWeathers}
\end{figure}

XiL (X-in-the-Loop) testing combines real-time simulation with physical or virtual components to validate systems in a controlled, repeatable environment. XiL enables early verification of algorithms, software, and hardware through closed-loop interaction with a simulated environment, reducing the need for full system prototypes.
XiL testing requires a high-definition map of the test environment, together with comprehensive sensor and weather models. These key components must be integrated into a real-time, deterministic simulation framework to ensure reproducible and repeatable results across multiple runs and weather conditions. 
This capability makes XiL particularly well suited for evaluating safety-critical and weather-dependent scenarios, such as those illustrated in Fig.~\ref{fig:XILWeathers}.

With the simulation loop and the weather and sensor models operating in real time, the simulated sensor data must be delivered to the perception stack (e.g., an object detection algorithm) through the same interfaces and drivers used by real-world sensors. This requires emulating the sensor's hardware and communication bindings to ensure realistic timing behavior and system interactions. Consequently, specialized software and, in some cases, dedicated hardware are needed to support this functionality.
Two common approaches are Direct Data Injection (DDI) and Over-the-Air (OTA) emulation. In OTA, a physical sensor observes outputs from an external emulator, thereby replicating real-world sensing conditions. DDI, in contrast, injects synthetic sensor data directly into the perception pipeline, bypassing the physical sensor. While DDI introduces additional integration complexity, it enables rapid, repeatable, and scalable testing across a wide range of scenarios without concerns related to sensor calibration. 
Regardless of the chosen approach, the quality of the emulation ultimately depends on the fidelity of the underlying models. Neither DDI nor OTA can compensate for an inaccurate sensor or environment model; high-quality models are essential for obtaining realistic and trustworthy results.

With the data correctly received by the Electronic Control Unit (ECU) as in the real world, the ECU must either be used or emulated to simulate the hardware constraints of the real-world vehicle, which has limited physical space, power, and computing resources. 
With the ECU in the simulation loop, it is possible to calculate and measure the time for each frame computation on the same hardware as the real-world counterpart in the car, thereby maintaining maximum hardware fidelity. 
Next, the ECU generates vehicle control commands, which are passed to the vehicle dynamics model. This model computes the vehicle's resulting motion and updates its position, orientation, and state. The simulation environment is then recomputed from the vehicle's new perspective, generating updated sensor outputs that are fed back to the ECU, thereby closing the loop.
The credibility of such an XiL framework depends on rigorous validation of each constituent component. This includes the environment model, sensor and noise models, data injection mechanisms, ECU implementation or emulation, and vehicle dynamics model. 
Only by validating these elements individually can confidence be established in the accuracy and reliability of the overall XiL system.

\textbf{Challenges.}
Several key challenges remain in XiL testing and validation.

\textbf{\textit{(i) Real-time performance and deterministic simulation:}} 
As sensor technologies continue to advance, simulation complexity increases significantly. Modern cameras now exceed 8 megapixels and offer 4K resolution; LiDAR sensors generate increasingly dense point clouds, with high-end systems producing over 10 million points per frame; and next-generation RADAR sensors incorporate elevation information, further increasing data dimensionality. Accurately modeling these sensors in simulation is computationally demanding, yet strict real-time performance must be maintained, as delays can adversely affect downstream perception, tracking, and control algorithms.
An additional challenge is maintaining temporal synchronization across all sensor modalities. Camera, LiDAR, RADAR, and other sensor outputs must be generated and delivered coherently within the same simulation cycle. Achieving real-time execution alone is insufficient; if simulating one sensor takes longer than the others, all sensor streams must still be synchronized and released simultaneously, rather than deferred to subsequent simulation cycles. Failure to preserve this temporal consistency can introduce artificial latency, corrupt sensor fusion processes, and lead to unrealistic system behavior and incorrect validation results.

\textbf{\textit{(ii) ECU integration and emulation fidelity:}} 
Fully integrating the ECU into the XiL simulation loop presents significant challenges, as it must operate under strict real-time constraints while accurately reproducing the behavior of the target hardware platform. Achieving this level of fidelity is non-trivial, particularly when emulating resource limitations such as computational performance, memory, communication latency, and power constraints. If the ECU emulation does not faithfully represent the real-world hardware, the resulting system behavior may deviate from reality, leading to inaccurate performance assessments and potentially misleading validation results.

\textbf{\textit{(iii)Vehicle dynamics model fidelity:}} 
Achieving a high-fidelity vehicle dynamics model requires extensive data collection and calibration across a wide range of operating conditions. Ideally, this process is conducted on a rolling road test bench, enabling systematic evaluation of all relevant combinations of engine speed, wheel speed, and load conditions. In addition, various sources of uncertainty, including measurement errors, parameter estimation inaccuracies, and model simplifications, can negatively affect simulation accuracy and closed-loop behaviour. These uncertainties must therefore be carefully quantified, validated, and mitigated to ensure credible and reliable XiL simulation results. Furthermore, winter conditions introduce large variations in tire-road friction and substantial uncertainty in vehicle and tire behavior, making it difficult to create representative XiL test scenarios. Consequently, reliable verification and validation of automated driving systems require high-fidelity weather-adaptive vehicle and tire models that can accurately capture the dynamics and uncertainties of snow- and ice-covered roads. However, developing such models is highly challenging.

\textbf{\textit{(iv) Validation methodologies:}} 
A lack of standardized credibility metrics for simulation models remains a significant challenge in XiL validation. Current regulations emphasize the need to assess and demonstrate the credibility of simulations, but do not prescribe a single, universally accepted evaluation framework or scoring methodology. This creates uncertainty in how simulation fidelity and trustworthiness should be quantified and compared across different platforms and approaches. As a result, establishing consistent and objective methods for assessing credibility remains an open research challenge that must be addressed to enable reliable benchmarking, regulatory acceptance, and validation of autonomous driving systems~\cite{unece_grva_2026_2,ECE2020,de-86/22}.

\textbf{Recommendation.}
For regulatory and standardization bodies such as the ECE, United Nations, and national authorities ~\cite{unece_grva_2026_2,ECE2020,de-86/22}, the methodology for assessing simulation credibility must be clearly defined and standardized to enable objective comparison across different XiL platforms and simulation approaches. Equally important is the establishment of guidelines for scenario selection, generation, and validation, ensuring consistent and representative evaluation of autonomous driving systems. Furthermore, the development of a common standard and unified description framework, or simulation language, is essential to promote interoperability, reproducibility, and comparability across tools, methodologies, and stakeholders.


%
%
\subsection{Transparency in perception and control}
\label{sec:transparency}

AV systems increasingly rely on Deep Learning (DL) to perform perception tasks. These DL-based perception modules transform high-dimensional, multimodal sensor data (camera, LiDAR, RADAR) into traffic scenario representations that are subsequently consumed by downstream planning and control components. While DL-based approaches outperform traditional systems in performance, they face a critical challenge: a lack of transparency in how they transform input data into scenario understanding. 
This opacity arises from the high complexity of DL models, which typically operate as black-box systems where internal decision-making processes are not interpretable to humans. In safety-critical applications, this limitation poses a fundamental obstacle to verification and validation (V\&V), in which explainability and traceability are required to provide evidence of compliance with safety requirements.

The challenge is further intensified by the broad range of ODD conditions in which AVs must operate safely, particularly adverse weather that degrades sensor data quality and reduces visibility. These factors significantly impair perception performance and increase uncertainty.

\textbf{Challenges.}
The key challenges can be summarized as follows: 
\textbf{\textit{(i) limited transparency}} in how DL-based perception models transform sensor inputs into scenario understanding,
\textbf{\textit{(ii) difficulty in assuring trustworthiness}} across DL component development and deployment pipelines,
\textbf{\textit{(iii) lack of a holistic approach}} to characterize and manage the inherent uncertainty of DL components, and
\textbf{\textit{(iv) high dimensionality}} and multimodal nature of sensor data, together with varying sensitivity to environmental conditions.

\begin{figure}[!b]
    \centering
    \includegraphics[width=1\linewidth]{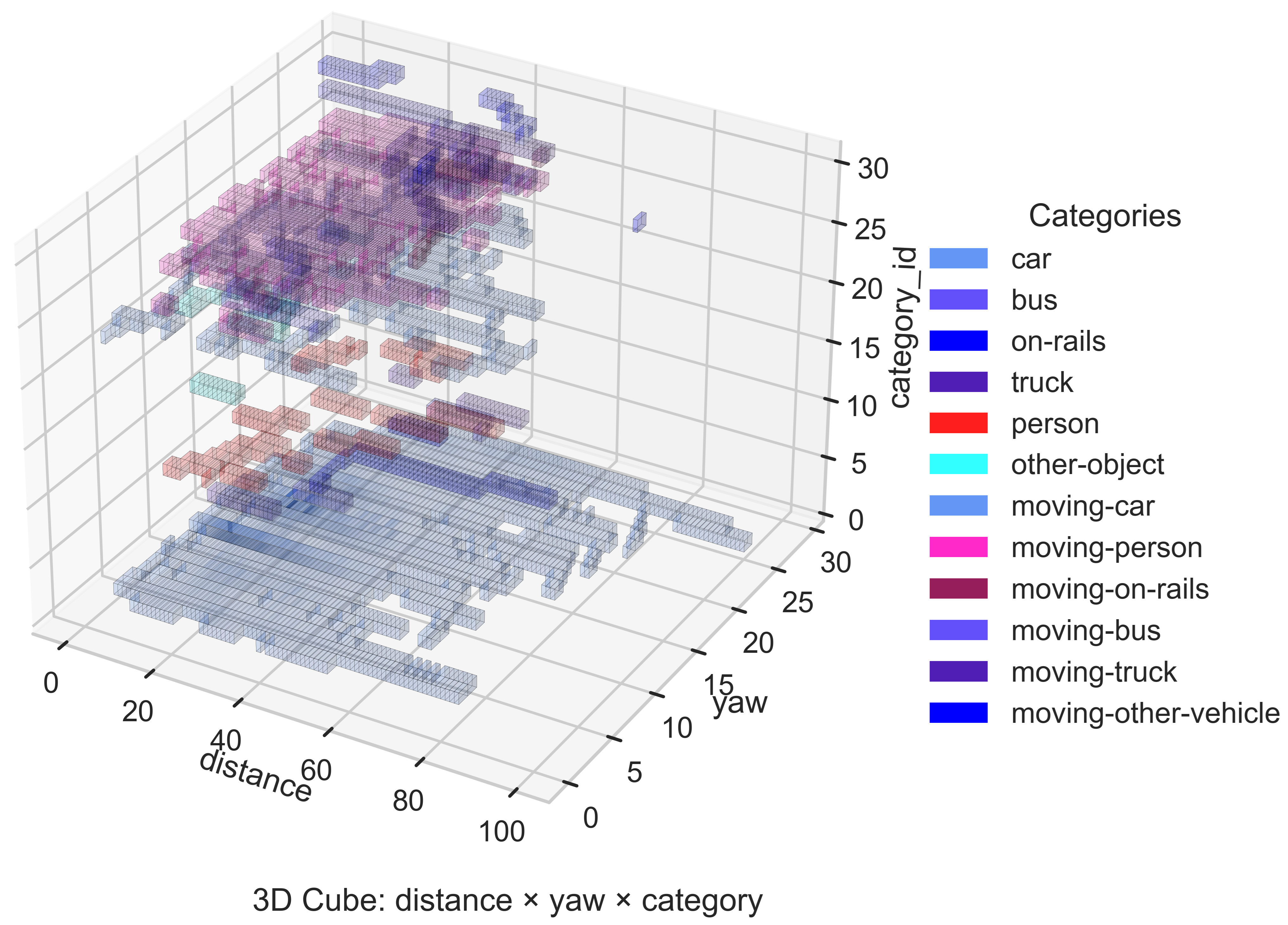}
    \caption{ROADVIEW Snowy Scenes dataset~\cite{snowyscenes} coverage analysis: 3D voxelized distribution of object annotations as a function of sensor distance, yaw angle, and object category.}
    \label{fig:fgi_3d_distance_yaw_category}
\end{figure}

\textbf{Recommendation.}
To achieve the required trustworthiness of DL-based perception systems, explainability should be integrated throughout the entire engineering lifecycle. Explainable AI (XAI) methods should be selected based on the objectives and stakeholders for each AI lifecycle phase, ensuring that human experts can understand, validate, and challenge model behavior. This integration improves transparency and provides evidence that perception outputs are sufficiently reliable for downstream planning and control functions. Such evidence is essential for safety assurance and compliance with emerging AI-related standards and regulations, including ISO 26262 \cite{ISO26262.2018}, ISO 21448 \cite{iso21448} (SOTIF), ISO/PAS 8800 \cite{iso8800}, and the EU AI Act\cite{AIAct}.

A key recommendation is to adopt a holistic uncertainty management framework that explicitly distinguishes between domain uncertainty, model epistemic uncertainty, and aleatoric uncertainty~\cite{BuiMarsh:ICHCAI2026}. 
Domain uncertainty originates from insufficient coverage of ODD conditions, critical scenarios, or combinations of these scenarios that form edge and corner cases.
Model epistemic uncertainty arises from limitations in model architecture and training, reflecting the gap between the implemented model and the theoretically optimal solution. Aleatoric uncertainty is the irreducible uncertainty inherent in the sensing and perception processes, arising from factors such as sensor noise, adverse weather conditions, occlusions, and annotation errors.

Managing uncertainty should begin during the AI development lifecycle. XAI methods should be applied throughout key phases, including data management, model learning management, and model inference management, especially via V\&V activities, to identify and reduce uncertainty wherever possible. Dataset representativeness should be systematically assessed against ODD and system requirements to identify coverage gaps and underrepresented scenarios that may contribute to domain uncertainty. Such analyses enable targeted data collection and safety-driven data augmentation to improve ODD coverage. Fig.~\ref{fig:fgi_3d_distance_yaw_category} illustrates an example of dataset coverage analysis, where the joint distribution of object categories, sensor distances, and object orientations is examined to assess the representativeness of the available data and identify potential coverage deficiencies.

While domain and model epistemic uncertainty are reduced during the development lifecycle, some uncertainty remains. Therefore, the system should provide uncertainty quantification mechanisms that estimate the residual uncertainty of its outputs. These uncertainty estimates provide important information for downstream tasks and support risk-aware decision-making. 

During vehicle operation, the residual uncertainty identified during development must be managed through runtime assurance mechanisms. Runtime safety architectures should therefore incorporate supervisory monitors leveraging anomaly detection and uncertainty quantification techniques. Together, these mechanisms enable the detection of conditions associated with elevated operational risk and support mitigation actions such as graceful degradation or minimal-risk maneuver. 

Finally, uncertainty estimates, detected anomalies, and other evidence generated by XAI should be systematically logged and linked to vehicle decisions to establish end-to-end traceability. Such traceability supports the safety argument that vehicle behavior remains consistent with safety goals under both nominal and degraded operating conditions, including adverse weather scenarios.

%
%
\subsection{Standardizations and Regulation}
\label{sec:stand_and_reg}

The large-scale deployment of weather-resilient AV perception and control systems is hindered by regulatory and policy barriers. Thus, policymakers (e.g., the EU and its Member States) must align fragmented Connected and Automated Vehicle (CAV) regulatory frameworks, clarify liability, and establish standardized testing and validation methodologies, particularly addressing adverse weather~\cite{EUcommission:vehicle2022}.

In the context of the ROADVIEW project, we have identified the following six key \textit{policy needs}:

\begin{enumerate} 
    \item \textbf{A harmonized framework} is needed to address fragmented CAV testing regulations and enable cross-border validation, particularly under harsh weather conditions. Legal constraints, combined with unpredictable weather, make comprehensive testing across various scenarios even more challenging.

    \item \textbf{A harmonized CAV liability framework} is needed to clarify responsibility in accidents, especially when following Vehicle-2-Everything (V2X) driving advice in harsh weather. Inconsistent regulations across EU Member States create legal uncertainties that could hinder the commercialization of weather-robust technologies across the European single market.
    
    \item \textbf{Balancing privacy and security} in CAV data collection is needed. Advanced AI-based technologies enhance CAV reliability in adverse weather conditions by using cameras. Yet, this may conflict with the General Data Protection Regulation (GDPR\footnote{\url{https://eur-lex.europa.eu/eli/reg/2016/679/oj/eng}}). Addressing this tradeoff is essential for enabling effective data sharing within the CCAM ecosystem while ensuring compliance and road safety.
    
    \item \textbf{Need for clear best practices and guidelines for AI use} in CAVs to ensure transparent, effective, and compliant decision-making under the AI Act \cite{AIAct}. AI-based innovations may be classified as high-risk, requiring significant effort to meet regulatory requirements and potentially impacting development and deployment.
    
    \item \textbf{An EU-wide regulatory hybrid methodology for CAV testing and approval} is required, with policymakers considering the metrics and test methodologies developed by various European projects. Integrating these into regulations could enhance simulation credibility, especially in harsh weather, and accelerate CAV deployment.
    
    \item \textbf{Need for mandatory EU/United Nations (UN) regulations following V2X communication standards} for adverse weather to ensure safety, interoperability, and large-scale adoption. Policymakers should consider new contributions within the European Telecommunication Standards Institute (ETSI\footnote{\url{https://www.etsi.org/}}) to strengthen regulatory frameworks and drive industry-wide compliance.
    
\end{enumerate}

In the context of ROADVIEW, the commercialization of new project technologies has encountered legal and regulatory barriers that could hinder future use of project results unless these barriers are addressed. Before examining the specific challenges faced by the ROADVIEW project, it is essential to understand the broader regulatory framework governing CAVs in the EU. Currently, CAV regulations in Europe remain fragmented, with responsibilities divided between the EU and its individual Member States. The regulatory landscape consists of \textit{three key dimensions}:
\begin{enumerate}
    \item \textbf{Vehicle design \& technical specifications:} This legal dimension sets standards for safety, efficiency, and operational performance. These aspects fall under the regulatory competence of the EU and the UN, ensuring harmonization across different jurisdictions. 
    \item \textbf{Traffic rules:} This dimension governs the movement of vehicles, pedestrians, and cyclists on roads to ensure safety and maintain order. These rules are determined at the national level, though they are often influenced by international agreements and conferences. 
    \item \textbf{Vehicle liability:} This last dimension defines responsibility when an automated system violates legal requirements. As liability laws are established at the national level, their interpretation varies across Member States, adding complexity to the legal landscape.
\end{enumerate}

At the national level, challenges often emerge due to the absence of a unified European approach to
CAV regulations. National challenges were analyzed within the ROADVIEW framework and grouped into \textit{two core categories: }
\begin{enumerate}
    \item \textbf{Data collection and testing: }Data collection and testing are essential for CAVs as they validate and refine prediction, perception, and decision making, enhancing safety, reliability, and the adaptability of machine learning/AI models to real-world conditions. However, in the European context, there is no unified approach to CAV data collection and testing. Each EU Member State sets its own regulations for obtaining permits to test CAVs on open roads, creating significant barriers to EU-wide testing and data collection. This challenge is further amplified when testing CAVs in harsh weather conditions, as such conditions can only be predicted, not guaranteed. A harmonized and simplified European framework could facilitate more efficient cross-border CAV testing, ultimately accelerating technological progress and deployment.
    \item \textbf{Liability of CAVs:} In the event of an accident, liability for human-driven vehicles rests with the driver. However, for CAVs, it remains unclear who would be responsible if a CAV followed V2X driving advice from an infrastructure system and caused an accident. Furthermore, liability regulations and their interpretations can vary significantly across EU Member States. The lack of harmonization and an EU-wide definition of liability is expected to slow down the commercialization of the ROADVIEW V2X Maneuver Coordination Services across the European single market.
\end{enumerate}

\textbf{Challenge.}  
Focusing on the EU and UN dimensions, the ROADVIEW consortium has identified potential barriers that could hinder the large-scale adoption of the proposed results:
\begin{enumerate}
\item \textbf{Privacy and data sharing:} In Europe, GDPR governs the collection, processing, storage, and sharing of personal data within the EU and the European Economic Area (EEA), imposing obligations on organizations to protect data security and transparency of users. Several ROADVIEW technologies rely on image acquisition through light and thermal cameras to guide CAVs safely and effectively, particularly in harsh weather conditions. These
technologies collect personal data, such as license plate numbers and images of road users, which are then shared within the CCAM ecosystem to enhance collective awareness of CAVs. Although various anonymization software solutions exist, achieving full GDPR compliance is not guaranteed, posing potential legal and ethical challenges. While ROADVIEW’s technology enhances CAV safety and efficiency, especially in harsh weather conditions, ensuring GDPR compliance remains a critical challenge.

\item \textbf{AI classification:} AI is essential for CAVs as it enables real-time perception, decision-making, and control, allowing vehicles to navigate safely without human intervention. AI is used to process sensor data, predict the behavior of other road users, and optimize routes. In Europe, the AI Act governs the development, deployment, and use of AI systems by categorizing them based on risk levels – unacceptable, high, limited, and minimal risk – to ensure safety, transparency, and compliance with fundamental rights while fostering innovation. Under the AI Act, innovations from the ROADVIEW project may be classified as high-risk, requiring technology manufacturers to make significant additional efforts to meet requirements for AI risk management, transparency, robustness, human oversight, and continuous monitoring. While AI is increasingly integrated into ADAS and CAV technologies, clear EU best practices and guidelines for ensuring efficient, transparent, and effective AI-driven decision-making and data management within CAV ecosystems are still lacking. 

\item \textbf{Hybrid validation methodology for CAV testing and approval:} Validation methodologies are structured approaches used to assess and ensure the safety, reliability, and performance of CAVs before they enter the market. In Europe, these methodologies are regulated by the (EU) 2022/1426\footnote{\url{https://eur-lex.europa.eu/eli/reg_impl/2022/1426/oj/eng}} and the UNECE Regulation 157\footnote{\url{https://op.europa.eu/en/publication-detail/-/publication/36fd3041-807a-11eb-9ac9-01aa75ed71a1}}. ROADVIEW developed a combination of virtual and real-world environments, a hybrid environment, to simulate CAVs’ behavior, using an XiL testing environment to accelerate the deployment of safe and robust automated driving functions, especially in harsh weather conditions. However, despite various norms and regulators emphasizing the benefits of this type of simulation environment, there is no clear
EU-wide regulatory methodology for validating the credibility of these simulations, inevitably delaying the time-to-market of innovations. 

\item \textbf{EU \& UN adoption of V2X communication standards for adverse weather:} Communication standards provide essential guidelines and specifications to ensure the safety, interoperability, and performance consistency of V2X communication across vehicle systems and infrastructure. These standards enhance collective awareness within the broader CCAM ecosystem. ROADVIEW is actively contributing to the development of V2X communication standards for adverse weather conditions within ETSI. While establishing such standards within the CAV industry marks progress towards Europe-wide implementation, compliance remains voluntary, as no mandatory EU or UN regulations currently enforce them,
inevitably affecting their pathway to large-scale adoption.
\end{enumerate}

\textbf{Recommendation.}
Consequently, there is a need for a harmonized European approach to tackle fragmented CAV regulations, ensure legal clarity on liability, and establish standardized testing methodologies. Additionally, clear guidelines on AI transparency, data privacy, and V2X communication standards are essential to facilitate large-scale adoption while maintaining public trust and compliance with EU regulations. 
To fully realize the benefits of ROADVIEW’s innovations, or any attempt toward weather-aware autonomy, policymakers must actively promote cross-border collaboration, invest in resilient transnational data-sharing infrastructure, and streamline regulatory frameworks. By taking these steps, Europe can reinforce its leadership in automated mobility and enable the successful deployment of weather-resilient CAVs – ultimately supporting the EU’s Vision Zero and Smart and Sustainable Mobility Strategies. 

%
%
\section{Conclusions}
\label{Sec:Conclusions}

Despite impressive advances in automated driving, achieving robust autonomy under severe weather conditions remains a major scientific, engineering, and regulatory challenge. Drawing on four years of development, experimentation, and large-scale demonstrations within the ROADVIEW project, this paper highlighted ten core challenges that continue to hinder reliable operation in snow, rain, fog, and other adverse environments. These challenges span the entire autonomous driving pipeline, from the availability of representative datasets and multimodal sensing, through weather-aware perception, localization, and decision-making, to simulation, validation, transparency, collective perception, and regulatory standardization. Our findings demonstrate that weather resilience cannot be achieved through isolated improvements in individual components but instead requires a holistic systems-level approach that tightly integrates sensing, AI, communication, simulation, and validation methodologies. Addressing these challenges will demand stronger collaboration between academia, industry, standardization bodies, and policymakers.
We believe that overcoming these barriers is essential for realizing the next generation of trustworthy, scalable, and weather-resilient automated vehicles, and ultimately for enabling the development of  \textit{snow-autonomous} vehicles.

It is important to note that these challenges and the recommendations derived from lessons learned in the ROADVIEW project are by no means unique to autonomous vehicles. 
Rather, they should be considered as general findings applicable across a wide range of autonomous robotic systems. 
Examples include search-and-rescue robots operating in smoke- or debris-filled environments, planetary-exploration robots operating under extreme lighting conditions, agricultural robots dealing with occlusion, and machines deployed in low-infrastructure settings. 
Ultimately, our goal is to advance robotic autonomy by developing robust perception and control systems that can adapt, recover, and learn under challenging real-world conditions.

%
%
\section{Acknowledgement}
\label{Sec:Acknowledgements}
This work was co-funded by the European Union. Views and opinions expressed are however those of the authors only and do not necessarily reflect those of the European Union or European Climate, Infrastructure and Environment Executive Agency (CINEA). Neither the European Union nor the granting authority can be held responsible for them. Project grant no. 101069576. 

UK participants in this project are co-funded by Innovate UK contract no.10045139. 

Swiss participants in this project are co-funded by the Swiss State Secretariat for Education, Research and Innovation (SERI) under contract no. 22.00123

\bibliographystyle{IEEEtran}
\bibliography{roadview} 
 
\end{document}